\documentclass[conference]{IEEEtran}
\usepackage{times}

\usepackage[numbers]{natbib}
\usepackage{multicol}
\usepackage[bookmarks=true]{hyperref}

\usepackage{graphicx}

\usepackage{xcolor}

\usepackage{amsmath}
\usepackage{amssymb}
\usepackage{booktabs}
\usepackage{tabularx}
\usepackage{xspace}

\usepackage{subcaption}

\DeclareMathOperator*{\argmin}{arg\,min}

\definecolor{MyDarkBlue}{rgb}{0,0.08,1}
\definecolor{MyDarkGreen}{rgb}{0.02,0.6,0.02}
\definecolor{MyDarkRed}{rgb}{0.8,0.02,0.02}
\definecolor{MyDarkOrange}{rgb}{0.40,0.2,0.02}
\definecolor{MyPurple}{RGB}{111,0,255}
\definecolor{MyRed}{rgb}{1.0,0.0,0.0}
\definecolor{MyGold}{rgb}{0.75,0.6,0.12}
\definecolor{MyDarkgray}{rgb}{0.66, 0.66, 0.66}
\definecolor{MyWineRed}{rgb}{0.694,0.071, 0.149}
\definecolor{nicegreen}{rgb}{0.1, 0.6, 0.2}
\makeatletter
\def\blfootnote{\gdef\@thefnmark{}\@footnotetext}
\makeatother

\newcommand{\model}{RoboCraft\xspace}

\begin{document}

\title{\model: Learning to See, Simulate, and Shape Elasto-Plastic Objects with Graph Networks}


\author{\authorblockN{Haochen Shi$^{*}$}
\authorblockA{Stanford University\\
hshi74@stanford.edu}
\and
\authorblockN{Huazhe Xu$^{*}$}
\authorblockA{Stanford University\\
huazhexu@stanford.edu}
\and
\authorblockN{Zhiao Huang}
\authorblockA{UC San Diego\\
z2huang@eng.ucsd.edu}
\and
\authorblockN{Yunzhu Li}
\authorblockA{MIT\\
liyunzhu@mit.edu}
\and
\authorblockN{Jiajun Wu}
\authorblockA{Stanford University\\
jiajunwu@cs.stanford.edu}
}



\maketitle

\IEEEpeerreviewmaketitle
\begin{abstract}

Modeling and manipulating elasto-plastic objects are essential capabilities for robots to perform complex industrial and household interaction tasks (e.g., stuffing dumplings, rolling sushi, and making pottery).
However, due to the high degree of freedom of elasto-plastic objects, significant challenges exist in virtually every aspect of the robotic manipulation pipeline, e.g., representing the states, modeling the dynamics, and synthesizing the control signals.
We propose to tackle these challenges by employing a particle-based representation for elasto-plastic objects in a model-based planning framework.
Our system, \model, only assumes access to raw RGBD visual observations. It transforms the sensing data into particles and learns a particle-based dynamics model using graph neural networks (GNNs) to capture the structure of the underlying system.
The learned model can then be coupled with model-predictive control (MPC) algorithms to plan the robot's behavior.
We show through experiments that with just 10 minutes of real-world robotic interaction data, our robot can learn a dynamics model that can be used to synthesize control signals to deform elasto-plastic objects into various target shapes, including shapes that the robot has never encountered before.
We perform systematic evaluations in both simulation and the real world to demonstrate the robot's manipulation capabilities and ability to generalize to a more complex action space, different tool shapes, and a mixture of motion modes.
We also conduct comparisons between \model and untrained human subjects controlling the gripper to manipulate deformable objects in both simulation and the real world. Our learned model-based planning framework is comparable to and sometimes better than human subjects on the tested tasks.
\footnote{Project page: \url{http://hxu.rocks/robocraft/}.}

\blfootnote{*Denotes equal contribution, random order.}
\end{abstract}


\section{Introduction}

    Effective manipulation of deformable objects is an essential skill for robots deployed in real-world industrial and household environments.
    However, due to deformable objects' high degrees of freedom (DoF) and consequent challenges in state estimation and dynamics modeling, manipulating deformable objects requires significant innovations beyond the typical robotic paradigm that focuses only on rigid objects.
    Recent advances show promising results in manipulating clothes~\citep{miller2012geometric, matas2018sim, wu2019learning, ha2021flingbot, yin2021modeling, fabric_vsf_2021} and ropes~\citep{yan2020learning, sundaresan2020learning}, yet the manipulation of objects with high plasticity, such as dough or plasticine, poses a unique set of challenges and is currently underexplored~\citep{matl2021deformable}, despite the ubiquity of such objects in household and industrial settings.
    In this paper, we investigate how to empower robots to model and manipulate elasto-plastic objects based on raw RGBD visual observations.
    
    The primary challenges of manipulating deformable objects stem from their high DoFs, partial observability, and complex non-linear local interactions.
    Learning dynamics models directly from high-dimensional sensory data offers a promising data-driven avenue for us to perform effective planning. For example, model-based reinforcement learning~(RL) algorithms have achieved great success in various planning and control tasks~\citep{schrittwieser2020mastering,nagabandi2020deep,manuelli2020keypoints}. However, when faced with elasto-plastic objects, these prior methods may fail due to a lack of explicit exploitation of the objects' structure.
    Another thread of works represents deformable objects using particles and employs graph neural networks~(GNNs) to model their dynamics~\citep{mrowca2018flexible,li2018learning,li2020visual,sanchez2020learning,shlomi2020graph}. They have shown great generalization results, demonstrating the benefits of explicit structured modeling.
    However, most of them require full-state information and a particle-based simulator to provide particle-to-particle correspondence between frames. Such strong supervision is difficult to obtain from raw sensory data, limiting their use in real-world applications.
    Hence, the natural question to ask here is: would it be possible to model the dynamics and manipulate elasto-plastic objects in the real world solely based on RGBD visual observations, without needing particle-to-particle temporal correspondence?  

\begin{figure}[t!]
	\includegraphics[width=\columnwidth]{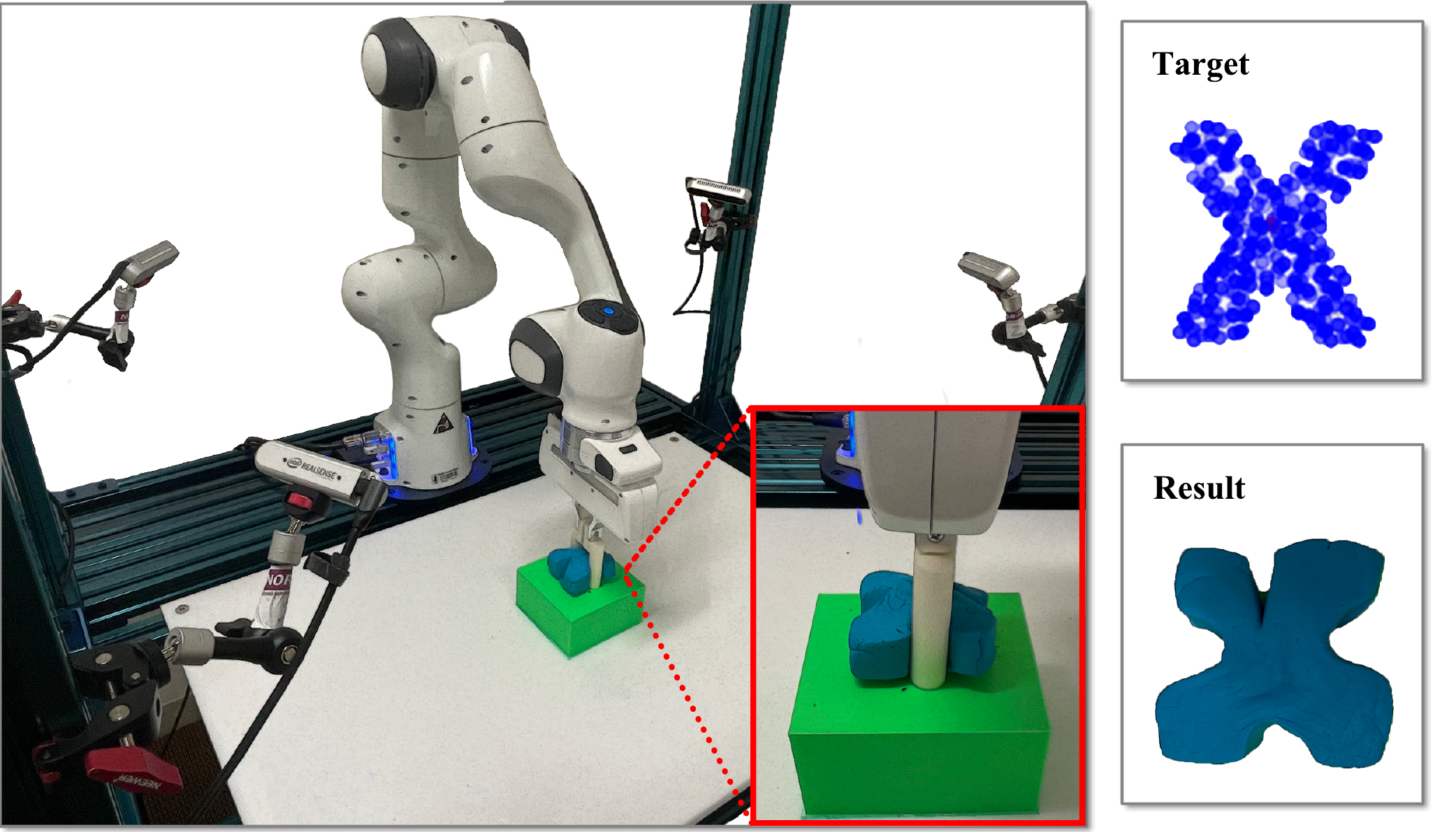}
	\caption{\textbf{\model}. The robot uses a parallel 2-finger gripper to shape an `X' conditioned on the target shape at the top right corner. The result is shown at the bottom right corner.}
	\label{fig:teaser}
\end{figure}
    
    To tackle this problem, we propose \model, a model-based planning framework that represents elasto-plastic objects using particles, but employs distribution-based loss functions and makes novel improvements over recently-developed GNNs to model the objects' dynamics.
    The learned dynamics model is then coupled with gradient-based trajectory optimization techniques to plan the robot's behaviors.
    The proposed approach closes the perception and control loops, which allows accurate modeling and manipulation of the elasto-plastic objects in both simulated and real-world settings. Specifically, our framework consists of (1) a perception module that constructs the particle representation of the object by sampling from the reconstructed object mesh, (2) a dynamics model that models the particle interactions using GNNs, and (3) a planning module that uses model-predictive control~(MPC) and solves the trajectory optimization problem using gradients from the learned model.
    Unlike prior learning-based particle dynamics works which assume temporal correspondence~\citep{mrowca2018flexible,li2018learning,li2020visual,sanchez2020learning,shlomi2020graph}, we train the dynamics model directly from raw visual data using loss functions that measure the distance between predicted and observed particle distributions.
    
In this work, we take concrete steps towards more general deformable object manipulation by enabling robots to model and manipulate elasto-plastic objects.
Through extensive evaluations in both simulation and the real world, we show that our model-based planning framework allows the robot (equipped with a parallel gripper) to deform the plasticine into complex target shapes (heart, alphabetical letters, etc.) purely based on raw visual inputs.
Notably, with just $10$ minutes of real-world interaction training data, our learned model can make accurate predictions of the object's deformation under planned actions, modeling the interactions between the gripper and the object, as well as the elastic/plastic deformation within the object.
We also asked amateur human subjects to control the robot to perform the same tasks, and our framework is on-par and sometimes more precise at achieving target shapes.

Our paper demonstrates that the \model framework can manipulate plasticine-like material accurately with a GNN model-based planning framework, and we hope the insights from this paper can inspire future work on more complex deformable object manipulation tasks.
\section{Related Work}

\subsection{Modeling the Dynamics of Deformable Objects }

Particle-based simulation is a popular family of methods to model deformable objects, wherein the dynamics are usually computed based on each particle's interaction with neighbor particles. Within this domain, position-based dynamics~\citep{muller2007position}, smoothed particle hydrodynamics~\citep{monaghan1992smoothed}, and material point methods~\citep{sulsky1995application} are widely used. Along with the advancement of particle-based methods, several differentiable simulators~\citep{hu2019difftaichi, holl2020learning, schoenholz2018end} that can propagate gradients through the model have been 
introduced to capture the dynamics of non-rigid bodies.

Data-driven approaches are another paradigm for learning physical dynamics~\citep{nagabandi2020deep, xu2018algorithmic}. Recently, people have demonstrated inspiring results on learning the dynamics of deformable objects such as clothes~\citep{lin2021learning}, ropes~\citep{chang2020model},  and fluid~\citep{legaard2021constructing}, with various representations including low-dimensional parameterized shapes~\citep{matl2021deformable}, keypoints~\citep{li2020causal}, latent vectors~\citep{kurutach2018learning}, and neural radiance fields~\citep{li20213d}. Inspired by prior approaches~\citep{li2018learning, sanchez2020learning, shlomi2020graph, battaglia2016interaction}, we choose to use a GNN-based method, due to its expressiveness in modeling the structure of an object, with very few assumptions on the underlying governing equations. Unlike prior methods that assume perfect perception and access to ground truth particle positions, our proposed method trains dynamics models directly from raw visual inputs, from which the one-to-one temporal correspondences among particles are hard to obtain.

\subsection{Manipulating Deformable Objects}

Deformable object manipulation is a long-standing challenge in robotics. However, many existing methods focus on objects such as ropes~\citep{wang2019learning, nair2017combining, yan2020self, sundaresan2020learning, lee2021sample, antonova2021bayesian, yan2020self}, cables~\citep{she2019cable, sanchez2020tethered, seita2020learning},  clothes~\citep{lin2020softgym, miller2012geometric, matas2018sim, wu2019learning, ha2021flingbot, antonova2021sequential, antonova2021dynamic}, and gauze~\citep{thananjeyan2017multilateral}. By contrast, we investigate the less-explored manipulation of elasto-plastic objects, such as  plasticine, which are only studied in limited previous works~\citep{sanchez2018robotic,li2018learning}.

As for manipulation of deformable elasto-plastic objects (dough, plasticine, etc), prior works propose to use either model-based~\citep{cretu2011soft, matl2021deformable, diffskill, contact} or adaptive methods~\citep{navarro2016automatic, cherubini2020model, yoshimoto2011active}. As mentioned earlier, it is possible to leverage simulations as models for actual manipulation~\citep{ganapathi2020learning}. However, the challenges in state and parameter estimation make it prohibitive to accurately simulate such objects and thus hinder the transfer of policies to the real world. Another popular approach is to learn from expert demonstrations~\citep{nadon2018multi}, which has been proved effective in shaping sand~\citep{cherubini2020model} and pizza dough~\citep{figueroa2016learning}. However, obtaining demonstrations is usually expensive.  \citet{matl2021deformable} propose to use soft robotic end-effectors to make a dough a sphere or rope-like shape with low-dimension representations such as bounding boxes. By contrast, \model equips the model with a particle representation and an expressive GNN, which enable our robot to shape objects to more complex unseen shapes such as hearts or alphabetical letters.  Our work is unique in the sense that we shape dough-like objects into complex shapes that contain semantic meanings.
\section{Method}
\begin{figure*}[t]
	\includegraphics[width=1.9\columnwidth]{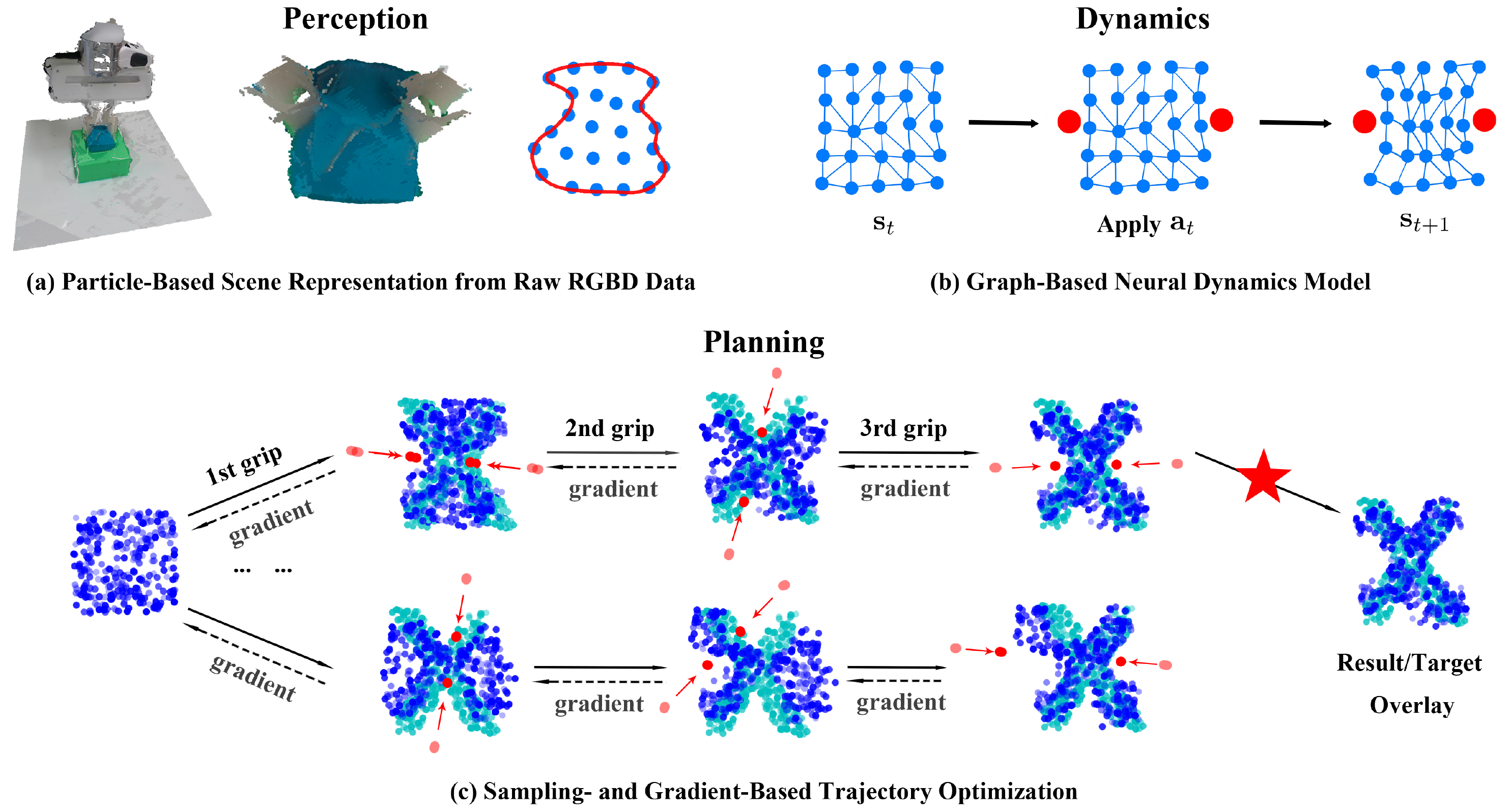}
	\caption{\textbf{Overview of \model.} (a) The perception module obtains the particle representation from RGBD cameras. The algorithm first crops out the object point cloud, then samples particles to represent the object. (b) The dynamics model predicts the object's deformation based on graph neural networks~(GNN). (c) After obtaining the learned dynamics model, we apply a combination of sampling- and gradient-based trajectory optimization techniques to solve the model-predictive planning problem.}
	\label{fig:overview}
\end{figure*}

\subsection{Problem Statement}
The objective of this work is to use a parallel 2-finger robot gripper to shape an elasto-plastic object to match a target shape $\mathbf{g}$. Specifically, we focus on using a sequence of pinching actions $\mathbf{a}_{0, ..., T-1}\in\mathcal{A}$, given an observation of the initial state $\mathbf{s}_0$ of the plasticine. At time step $\textit{t}$, the robot applies action $\mathbf{a}_t \in \mathcal{A}$ upon the plasticine, and the state of the plasticine transitions from $\mathbf{s}_t$ to $\mathbf{s}_{t+1}$ in response.

To predict the complex dynamics of the deformable plasticine, we propose to use a graph neural network~(GNN) $\Phi$ to learn the transition function $\Phi: \mathcal{S} \times \mathcal{A} \rightarrow \mathcal{S}$. This dynamics model takes as input environment observations $\mathbf{s}_{t-h,\dots,t} \in \mathcal{S}$ and actions $\mathbf{a}_{t-h,\dots, t} \in \mathcal{A}$ and predicts a future observation $\hat{\mathbf{s}}_{t+1}$, where $h$ is the length of history and $t$ is the current time step. 
With the learned dynamics model in hand, we can naturally formulate the manipulation task as a model-predictive control~(MPC) problem. The cost function $\mathcal{J}$ of the MPC problem measures the distance between the state of the plasticine at the last time step $T$ and the target shape $\mathbf{g}$. The choice of cost function $\mathcal{J}$ will be discussed in section \ref{sec: loss}. And a sequence of actions of length $\textit{T}$ can be selected by minimizing the cost function:
\begin{align}
    (\mathbf{a}_0, ..., \mathbf{a}_{T-1}) = \argmin_{\mathbf{a}_{0,\dots ,T-1} \in \mathcal{A}} \mathcal{J}(\Phi(\mathbf{s}_0, (\mathbf{a}_0, ..., \mathbf{a}_{T-1})), \mathbf{g}) 
\end{align}
Figure \ref{fig:overview} shows the overall framework of \model.

\subsection{Particle Sampling from Raw RGBD Data}
\begin{figure}[t]
	\includegraphics[width=\columnwidth]{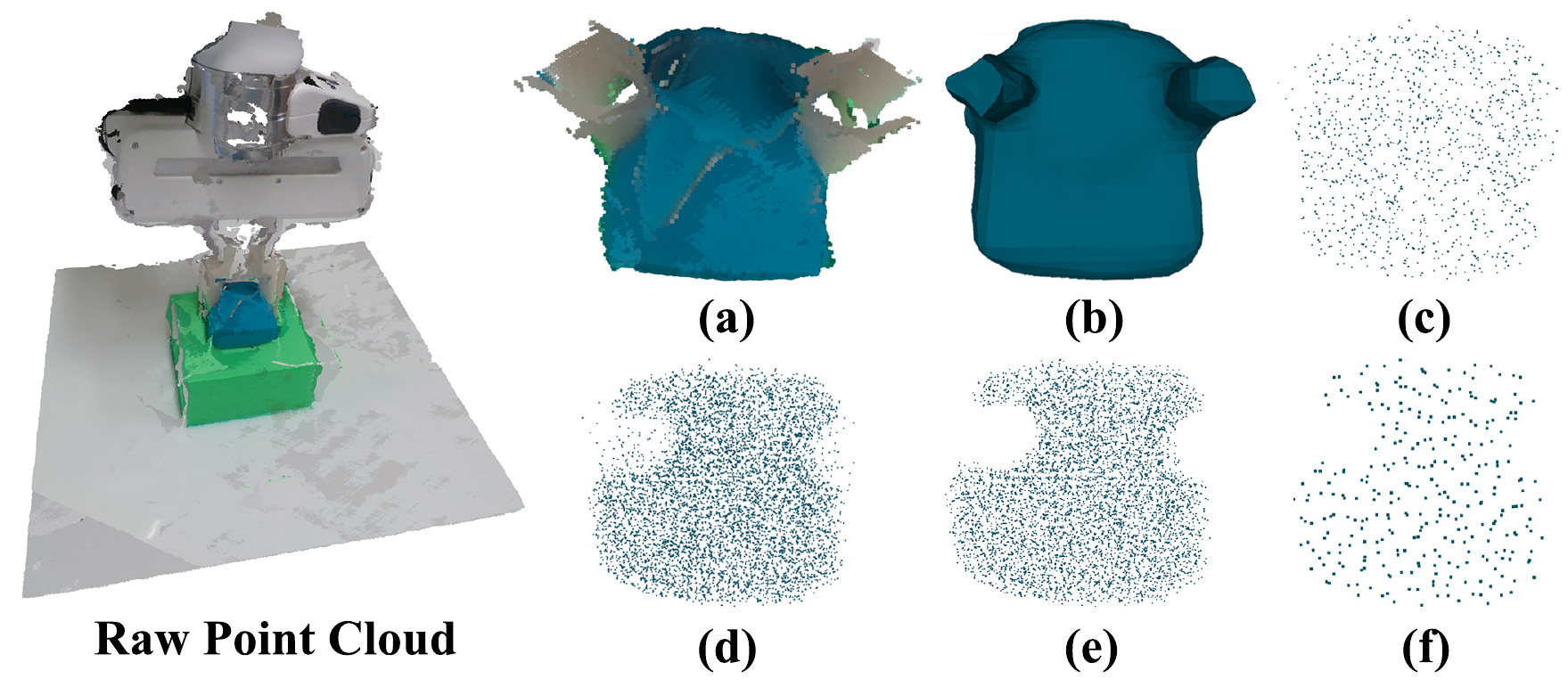}
	\caption{\textbf{The particle sampling procedure.} Left: Visualized point cloud for the manipulation scene. Right: (a) Extracted point cloud for the object and the gripper. (b) Reconstructed mesh surface. (c) Sampled particles within the mesh. (d) Refinement with physics prior. (e) Refinement with shape prior. (f) GNN-ready downsampled particles. }
	\label{fig:sampling}
\end{figure}

We aim to sample particles from the raw visual data (i.e., RGBD images) for the GNN-based dynamics model as shown in Figure~\ref{fig:overview}(a). 
To obtain useful particles, we need to sample particles that can represent the shape of the object from heavily occluded raw inputs. We illustrate how we obtain the required particles through preprocessing, surface reconstruction, refinement with physics prior, refinement with shape prior, and postprocessing as shown in Figure~\ref{fig:sampling}.

\subsubsection{Preprocessing}
We first convert RGBD images into point clouds, using the intrinsic and extrinsic parameters of the cameras.
We then use RANSAC~\citep{fischler1981random} to detect the plane with the largest support in the raw point cloud (i.e., the platform on which we place the object) and remove the inliers of this plane. We also remove other noisy backgrounds based on the location of the object. The outcome is shown in Figure~\ref{fig:sampling}(a).

\subsubsection{Surface Reconstruction}
Two popular surface reconstruction algorithms are alpha shapes~\citep{edelsbrunner1994three} and ball pivoting~\citep{bernardini1999ball}. However, these algorithms are not robust to occlusion. We instead use Poisson surface reconstruction~\citep{kazhdan2006poisson}, a method that can smooth the occluded surface by solving a regularized optimization problem. 
To prepare for Poisson surface reconstruction, we estimate the normal for each point in the point cloud by calculating the principal axis of its adjacent points.  The outcome is shown in Figure~\ref{fig:sampling}(b).

\subsubsection{Refinement with Physics Prior} 
We exploit the knowledge that when the gripper is not touching the plastic object, the shape of the object should largely remain unchanged. Hence, after sampling the particles as in Figure~\ref{fig:sampling}(c), missing particles that are not reconstructed from the previous step (e.g., an artifact caused by shadow) are patched using points from the previous frame. The outcome is shown in Figure~\ref{fig:sampling}(d).

\subsubsection{Refinement with Shape Prior}
In the process of gripping, the object is occluded by itself as well as the gripper. The heaviest occlusion occurs when the gripper is pinching the object. We first perform a simple color-based filter to extract the object point cloud. However, the point cloud of the plasticine is incomplete due to occlusion, and thus reconstructing a watertight mesh from it becomes difficult. We propose leveraging shape prior to refine the point cloud. Specifically, after obtaining a smooth and watertight mesh from the full point cloud (object and gripper), we then sample particles inside the reconstructed mesh based on its signed distance function~(SDF). We finally use the fingers' SDFs to remove the points inside them. The outcome is shown in Figure~\ref{fig:sampling}(e).

\subsubsection{Post-processing}
To further improve the quality of the sampled point cloud, we apply voxel down-sampling to get a uniform point distribution and remove the points that are statistical outliers.
We then use the farthest point sampling method~\citep{moenning2003fast} to further down-sample the point cloud of the plasticine into a reasonable number of points for the GNN. The final output is shown in Figure~\ref{fig:sampling}(f).

\subsection{Learning Object Dynamics via Graph Networks}
\label{sec:gnn}
We now introduce how \model constructs a particle graph and uses graph neural networks to model the dynamics of the system, as shown in Figure \ref{fig:overview}(b). 

\subsubsection{Graph Building} In a graph formed by states $\mathbf{s}_t=\left(\mathcal{O}_t, \mathcal{E}_t\right)$, the vertices $\mathcal{O}_t$ of the graph are the particles $\mathbf{o}_{i,t}$ of the object. Specifically,  $\mathbf{o}_{i,t}=\left<\mathbf{x}_{i,t}, \mathbf{c}_{i,t}^o\right>$, where $\mathbf{x}_{i,t}$ is the position of particle $i$ at time $t$, and $\mathbf{c}_{i,t}^o$ denotes the corresponding attributes (e.g., mass, radius). The edges $\mathcal{E}_t$ between the vertices are computed dynamically over time from their spatial relationship. We connect all the neighbors within a predefined distance. For relations between particles (e.g., a pair of particles with an edge), we use $\mathbf{e}_k=\left<u_k, v_k, \mathbf{c}_k^e\right>$, where $1 \leq u_k, v_k \leq |\mathcal{O}_t|$ are the receiver particle index and sender particle index respectively, $k$ is the edge index, and $\mathbf{c}_k^e$ is the type of relationship (e.g., object internal relation or gripper-to-object relation).  

\subsubsection{Model Training} The goal of the GNN is to infer the system dynamics and predict the future from the current graph as $\hat{\mathbf{s}}_{t+1}=\Phi(\mathbf{s}_t, \mathbf{a}_t)$. We use $f_O^{\text{enc}}$ and $f_E^{\text{enc}}$  to encode the object features and the relation features respectively as follows: 
\begin{align}
h_{i,t}^o&=f_O^{\text{enc}}(\mathbf{o}_{i,t})\\ h_{k,t}^e&=f_E^{\text{enc}}(\mathbf{o}_{u_k, t}, \mathbf{o}_{v_k, t}, \mathbf{c}_k^e)
\end{align}
We then use an object function $f_O^{\text{dec}}$ and a relation function $f_E^{\text{dec}}$ to model the dynamics. The future state at time $t+1$ is predicted as 
\begin{align}
b_{k, t}&=f_{E}^{\text{dec}}(h_{k,t}^e)_{k=1,\cdots,|\mathcal{E}_t|}\\ 
\hat{\mathbf{o}}_{i, t+1}&=f_O^{\text{dec}}(h_{i,t}^o, \sum_{k \in \mathcal{N}_i} b_{k,t})_{i=1,\cdots,|\mathcal{O}_t|}
\end{align}
where $\mathcal{N}_i$ is a set of relations with particle $i$ as the receiver. During training, we also use multi-step message passing to handle instantaneous propagation of forces.

\subsection{Loss Functions}
\label{sec: loss}
Since our training data comes from sampled point cloud data, there is no one-to-one correspondence among the points of each frame as required by particle-wise losses.  We explore two loss functions to measure the similarity between the distributions of point cloud data.

\textbf{Earth mover’s distance (EMD).} Consider $\mathcal{O}_1, \mathcal{O}_2 \subseteq \mathbb{R}^3$. The EMD between them is defined as
\begin{align*}
\mathcal{L}_{\text{EMD}}(\mathcal{O}_1, \mathcal{O}_2) = \min_{\mu:\mathcal{O}_1 \rightarrow \mathcal{O}_2} \sum_{\mathbf{x} \in \mathcal{O}_1} \|\mathbf{x}-\mu(\mathbf{x})\|_2,
\end{align*} 
where $\mu: \mathcal{O}_1 \rightarrow \mathcal{O}_2$ is a bijection~\citep{rubner1998metric}. The EMD solves an assignment problem.
For all but a zero-measure subset of point set pairs, the optimal bijection $\mu$ is unique and invariant under the infinitesimal movement of the points.
Intuitively, in this work,  EMD matches distributions while preventing outliers in the point cloud by the bijection definition.

\textbf{Chamfer distance (CD).} 
The CD between $\mathcal{O}_1, \mathcal{O}_2 \subseteq \mathbb{R}^3$ is
\begin{align*}
\mathcal{L}_{\text{CD}}(\mathcal{O}_1, \mathcal{O}_2) =  \sum_{\mathbf{x} \in \mathcal{O}_1} \min_{\mathbf{y} \in \mathcal{O}_2} {\Vert \mathbf{x} - \mathbf{y} \Vert}_2^2 + \sum_{\mathbf{y} \in \mathcal{O}_2} \min_{\mathbf{x} \in \mathcal{O}_1} {\Vert \mathbf{x} - \mathbf{y} \Vert}_2^2.
\end{align*}

In the strict sense, we slightly abuse the term Chamfer distance because it does not satisfy the triangle inequality ~\cite{fan2017point}. Intuitively, CD finds the nearest neighbor for a point in the other set and sums the squared distances.

Our loss function is a weighted sum of the two distance functions mentioned above: $
\mathcal{L}(\mathcal{O}_1, \mathcal{O}_2) = \textit{w}_{1} \mathcal{L}_{\text{EMD}}(\mathcal{O}_1, \mathcal{O}_2) + \textit{w}_2 \mathcal{L}_{\text{CD}}(\mathcal{O}_1, \mathcal{O}_2)$.
Based on experimental results, we find the optimal hyper-parameters are $\textit{w}_1=0.9, \textit{w}_2=0.1$.

\subsection{Model Predictive Control}
In this section, we introduce a new model predictive control~(MPC) algorithm for controlling the actuators to manipulate the plasticine as shown in Figure~\ref{fig:overview}(c).

\subsubsection{Action Space}
We simplify the action space of each grip into a parameterized space: $\{\textit{x}, \textit{y}, \textit{z}, \textit{r}_z, \textit{d}\}$. In this simplified space, $\{\textit{x}, \textit{y}, \textit{z}\}$ are bounded locations indicating the center of the grip, i.e. the midpoint of the line segment connecting the centers of mass of the gripper's two fingers. $r_z$ is the robot gripper's rotation about the $z$ axis (i.e., the vertical axis). The rotations about the $x$ and $y$ axes are always 0. $\textit{d}$ represents the minimal distance between gripper fingers during this pinch. The robot gripper pinches the plasticine at a constant speed. 

\subsubsection{Goal-Conditioned MPC} We denote $\mathbf{g}$ as the target shape of the plasticine and $\mathbf{a}_{0:T-1}$ as the action sequence sampled from the simplified action space, where $T$ is the time horizon. We denote the resulting trajectory after applying the control inputs as $\mathbf{s}_{0:T}$. The task is to determine the actions that minimize the distance between the actual outcome and the specified goal~$\mathcal{J}(\mathbf{s}_T, \mathbf{g}) \triangleq \mathcal{L}(\mathcal{O}_T, \mathcal{O}_\mathbf{g})$. 

We use gradient-based trajectory optimization to obtain the trajectory with the lowest cost as shown in Figure~\ref{fig:overview}(c). We first do random shooting in the simplified action space and compute the costs using the GNN-based dynamics model. We then apply the limited-memory BFGS~\citep{fletcher2013practical} algorithm on the lowest-cost trajectories to optimize the actions with gradients, where the loss is the same as our dynamics model's training loss.

In the standard form of MPC, the controller can use intermediate states as feedback to correct future actions. As a trade-off in this work, acquiring intermediate states with particle sampling would improve the effectiveness but reduce the time efficiency. Hence, we provide intermediate states from cameras only after each grip, ignoring the visual feedback within the duration of a grip. 

While each video in the training dataset includes three grips, we may require more or fewer grips when aiming at unseen shapes. To address this issue, we perform MPC with a fixed look-ahead horizon and execute one grip at a time. The robot performs additional grips until the loss is small enough or the maximum number of grips is reached.

\section{Experimental Setup}
\begin{figure}[t]
    \centering
	\includegraphics[width=0.9\columnwidth]{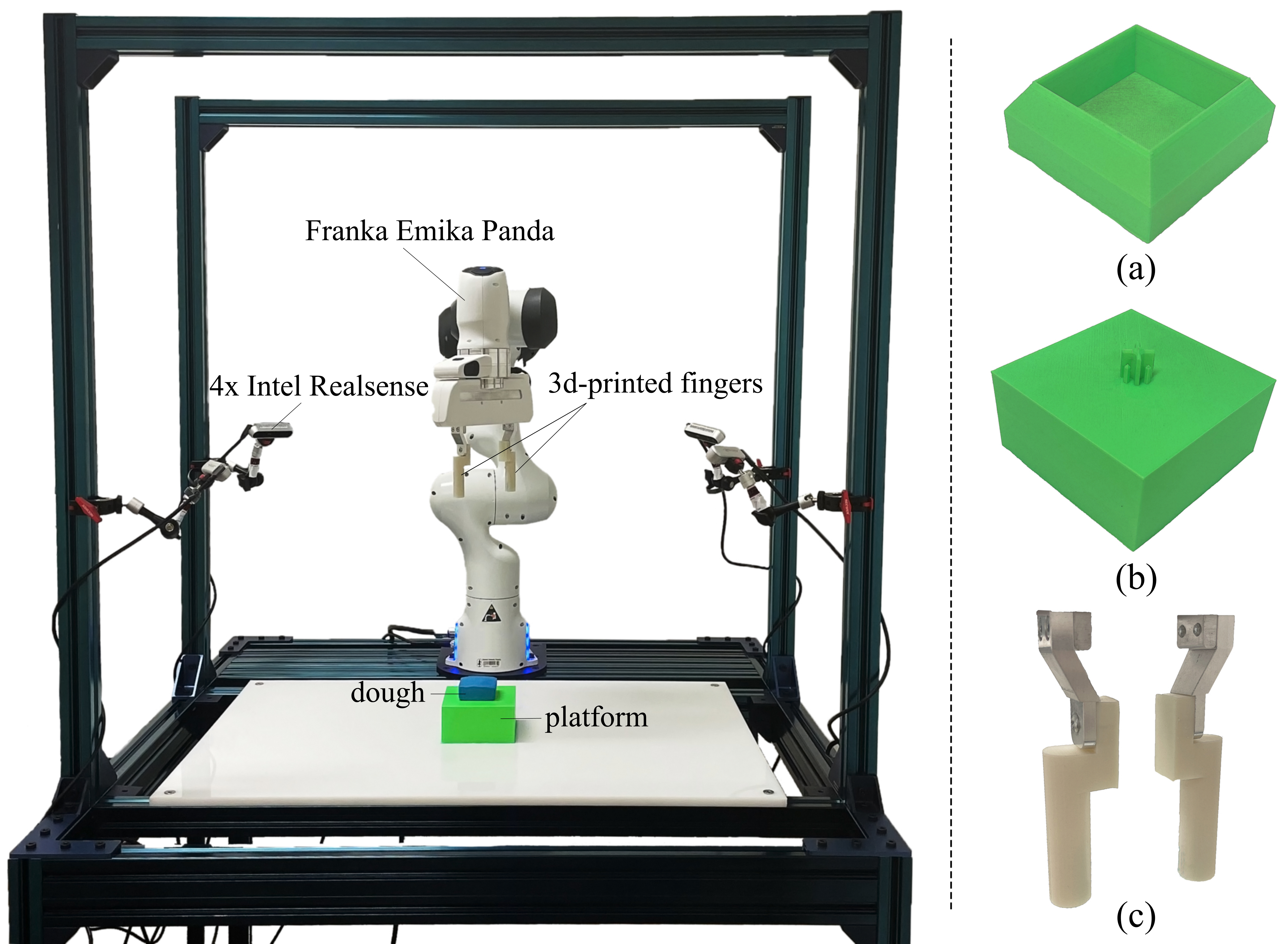}
	\caption{\textbf{Robot Setup.} Left: overview of the robot setup; Right: 3D-printed tools including (a) the mold for resetting, (b) the supporting platform, and (c) the parallel 2-finger gripper. }
	\label{fig:robot}
\end{figure}

\subsection{Physical Setup}
In Figure \ref{fig:robot}, we show the general setup in the simulator and real world. For real-world manipulation, we use a Franka Emika Panda robot arm with 7 DoF and Franka's parallel jaw gripper. We substitute the original grippers with a pair of 3D-printed cylindrical parallel gripper fingers (Figure~\ref{fig:robot}(c)). Four RealSense D415 RGBD cameras are fixed at four locations surrounding the plasticine to capture the RGBD images at 30Hz and 848$\times$480 resolution. The four cameras are calibrated to get the relative positions with respect to each other and the robot base to reconstruct the object geometry. We use a blue Play-Doh modeling compound as the deformable object. The initial shape of the plasticine is roughly a 6cm$\times$6cm$\times$2.5cm cuboid. In the coordinate frame where the robot base is at the origin, the initial position of the center of mass of the plasticine is at (43.00cm, 0.00cm, 7.25cm) on a 3D-printed platform, with a central protruding plastic rod to fix the object  (Figure~\ref{fig:robot}(b)). At the beginning of each episode, we stuff the plasticine into a 3D printed mold in the shape of a cuboid cavity (Figure~\ref{fig:robot}(a)) to reset the shape of the object. We will release STL files for all the 3D-printed parts.

\subsection{Data Collection}
We collect 6{,}000 frames (10 mins) of training data, including 50 episodes with a horizon of 120 frames. For each episode, three grips are applied to the plasticine. At the beginning of each episode, we manually shape the plasticine into a cuboid with a mold and plastic wrap. We knead the manipulated object before molding it to avoid wrinkles on the plasticine. The data collection behavior policy randomly selects parameters in the action space consisting of rotation angles, translations, and gripper fingertip distances. During each episode, we save the generated partial point clouds from each of the RGBD cameras and the robot joint poses from the robot controller. The data is only saved when the gripper is on the same plane as the object to optimize memory usage. 

\subsection{Tasks and Benchmarks}
The main challenges in deformable object manipulation are 1) the high DoF of the object, unlike the other deformable objects, and 2) a mixture of rigid and non-rigid motion modes. To highlight these challenges, we propose a series of tasks of shaping fixed and movable elasto-plastic objects, which require rich local operations in addition to global pose matching. With the final goal of real-world manipulation in mind, we first set up experiments in simulation~\citep{huang2021plasticinelab} with two capsule-shaped fingers and cuboid-shaped plasticine. The following is a brief overview of the tasks:
    
    \noindent \textbf{Shaping w/ Rigid Motion}: Grip a freely movable deformable object into the target shape. Gripper rotation and translation are allowed only in the plane parallel to the tabletop.
    
    \noindent \textbf{Shaping w/ 7-DoF gripper}: Grip a fixed deformable object into the target shape. There is no constraint of the parallel plane for the gripper - it can rotate freely in the 3D space.
    
    \noindent \textbf{Tool Selection}: Grip a fixed deformable object. For each grip, the agent can choose between two tools of different sizes. 
    
Along with these tasks, we propose a benchmark for deformable object shaping in both simulation and the real world. We use 26 alphabetical letters and five other shapes as our target shapes in the simulator. For the real robot, we use seven letters and five other shapes. Note that for letters with holes such as `B' or `P', we only focus on the contour because the current tools are not suitable for carving out the hollow portions. We believe that the proposed shapes span a large spectrum of possible shapes and can serve as a benchmark for future research works. 

\subsection{Evaluation Metrics}
We quantitatively evaluate each step of our whole framework including the particle sampling, the dynamics model training, and the manipulation results. We use Chamfer distance~(CD) and Earth mover's distance~(EMD) as our main metrics. In addition, we use Hausdorff distance (HD)~\citep{huttenlocher1993comparing} to measure whether every point of either set is close to a point of the other set.

\begin{figure}[t!]
    \centering
	\includegraphics[width=\columnwidth]{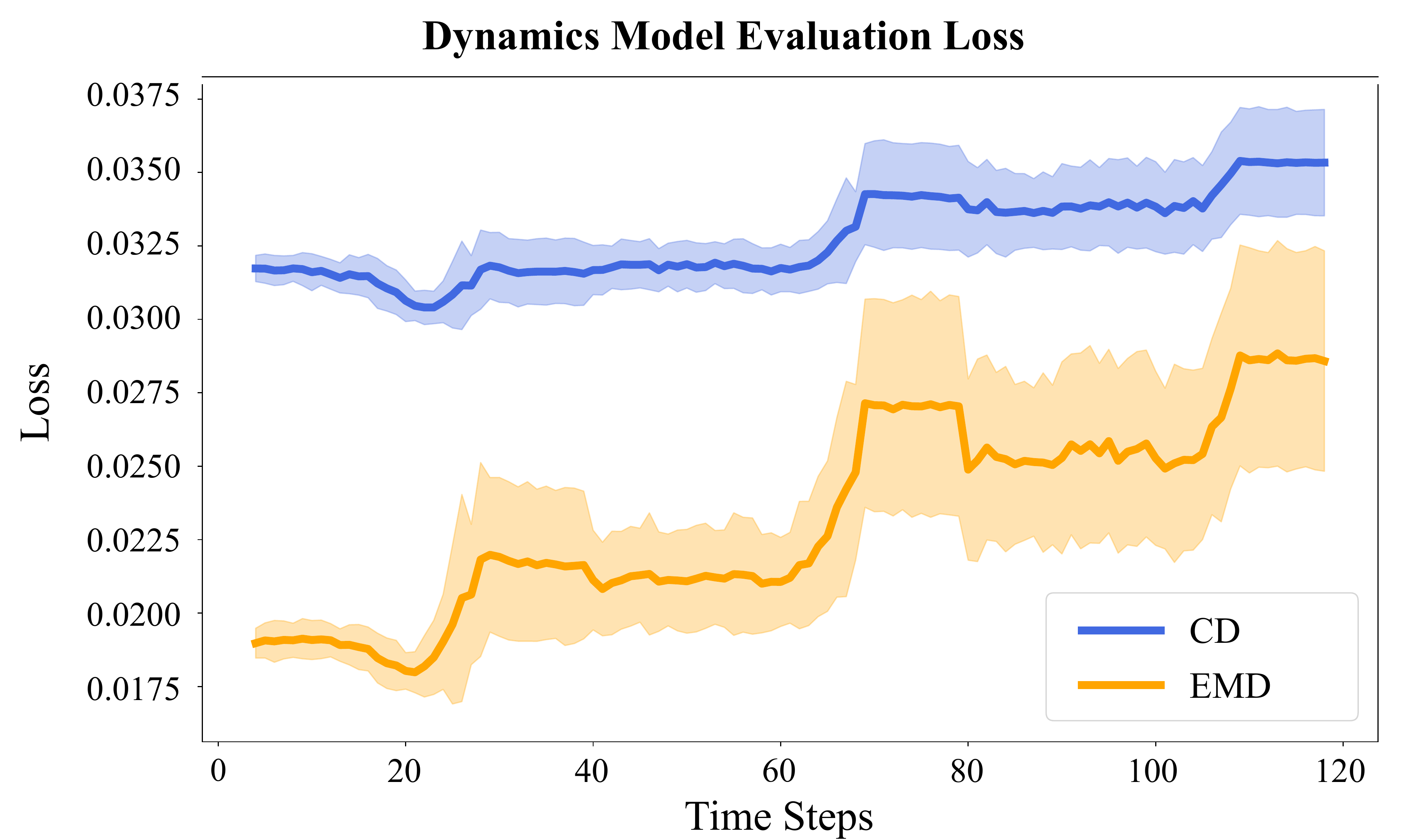}
	\caption{\textbf{Quantitative results on dynamics prediction.} The figure shows the CD and EMD between the predicted particles and the ground truth particles over time. The shaded area is one standard deviation.}
	\label{fig:eval_loss}
\end{figure}

\begin{figure}[t]
	\includegraphics[width=\columnwidth]{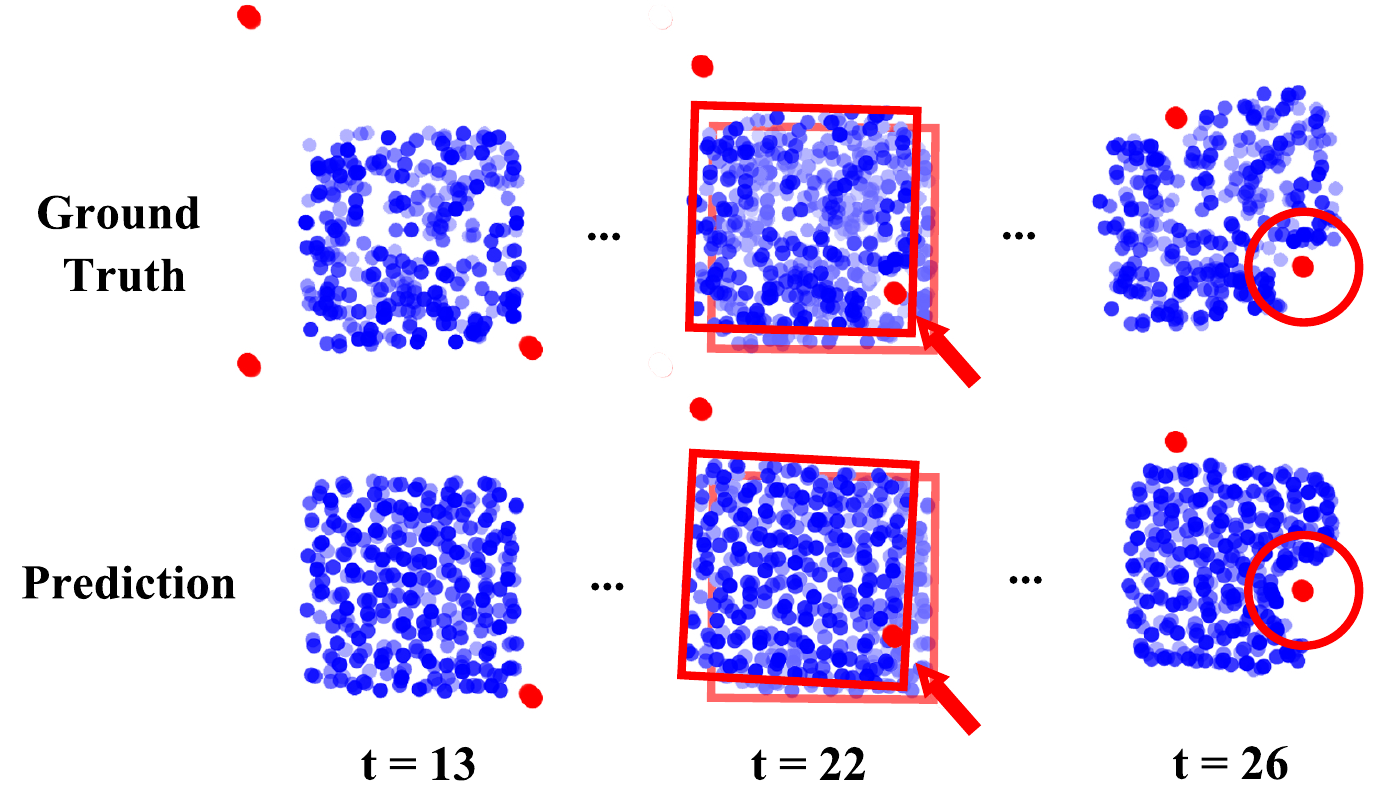}
	\caption{\textbf{Rigid motion and non-rigid motion over time frames.} The first row is the ground truth and the second row is the predicted motion from the dynamics model.}
	\label{fig:rigid-motion}
\end{figure}

\section{Experimental Results}

In this section, we evaluate the proposed method for various complex deformable elasto-plastic object manipulation tasks. We compare the proposed method with a set of baseline algorithms and untrained human subjects. We also carefully investigate various design decisions to determine the important factors for future researchers' reference. Lastly, we apply the lessons learned from simulation experiments to enable a robot to manipulate elasto-plastic objects in the real world.

\subsection{Results in Simulation}
\subsubsection{Sampling Results}

We first compare the performance of the proposed sampling method and the patch-based baseline. The patch-based method extracts the partial point cloud of the plasticine based on color. It then reconstructs a convex hull around the incomplete point cloud and extracts points from the gripper point cloud within the reconstructed convex hull to patch the point cloud of plasticine.

In Table~\ref{tab: sampling}, we compute the average distance between the sampled particles and the ground-truth particles provided by the simulator. We find that our method (`crop-based') achieves lower losses for both CD and EMD and outperforms the patch-based baseline. These results resonate with the intuition that additional physics and shape priors of the gripper can significantly improve the sampling quality in occluded scenes.

\subsubsection{Dynamics Model Learning}

\begin{table}[t] 
\caption{Sampling results averaged over 120 frames in Chamfer distance (CD) and Earth mover’s distance (EMD).}
\centering
\begin{tabular}{lcc}
\toprule
Methods         & CD$\downarrow$ & EMD$\downarrow$ \\ 
\midrule
Patch-Based        & 0.0384 $\pm$ 0.003  & 0.0317 $\pm$ 0.005 \\
Crop-Based        & \textbf{0.0374} $\pm$ 0.001  & \textbf{0.0308} $\pm$ 0.002 \\
\bottomrule
\end{tabular}
\label{tab: sampling}
\end{table}

\begin{table}[t] 
\caption{The mean and standard deviation of the dynamics model's performance over 12{,}000 frames when trained with different loss functions.}
\centering
\begin{tabular}{lccc}
\toprule
Loss Types         & CD$\downarrow$ & EMD$\downarrow$ & HD $\downarrow$ \\ 
\midrule
CD Loss        & 0.0344 $\pm$ 0.003  & 0.0267 $\pm$ 0.006 &  0.0851 $\pm$ 0.018\\
EMD Loss     & 0.0328 $\pm$ 0.002  & \textbf{0.0230} $\pm$ 0.004 & 0.1068 $\pm$ 0.011\\
Mixed Loss    &  \textbf{0.0327} $\pm$ 0.002  & 0.0231 $\pm$ 0.004  &  \textbf{0.0790} $\pm$ 0.016 \\
\bottomrule
\end{tabular}
\label{tab: dynamicsmodel}
\end{table}

\begin{figure}[t!]
    \centering
    \includegraphics[width=\columnwidth]{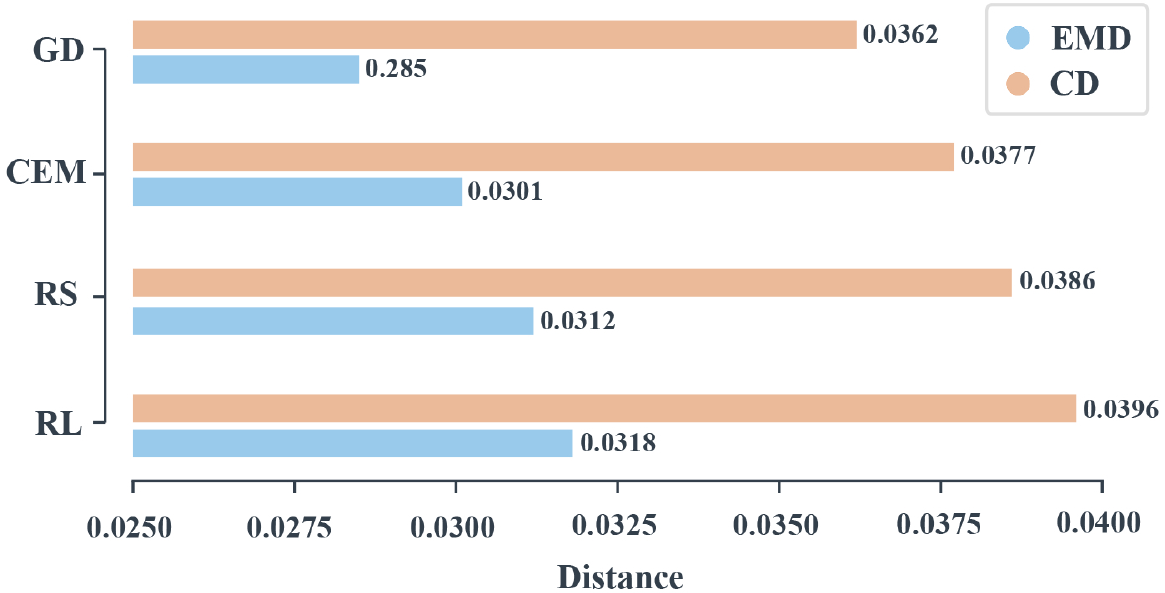}
	\caption{\textbf{Comparison of different trajectory optimization methods.} We benchmarked four different optimization algorithms, including gradient descent~(GD), cross-entropy Method~(CEM), random shooting~(RS), and reinforcement learning~(RL). CD and EMD are computed on the target shape `A'. Lower is better.}
	\label{fig:opt}
\end{figure}	

We train the GNN model as explained in Section~\ref{sec:gnn}. Specifically, we first construct the graph by connecting edges between two vertices with the proximity threshold $d=0.05$. For each edge and vertex, we encode them with $3$-layer MLPs with hidden and output layers each of size $150$. The propagator consists of one fully connected layer with an output layer size of $150$. The motion predictor is another $3$-layer MLP with a hidden layer size of $150$. All of the neural networks use ReLU activations. The models are trained for $100$ epochs with the Adam optimizer~\citep{kingma2015adam}, a batch size of $4$, and a learning rate of 1e-4.

We then evaluate GNN-based dynamics models trained with different loss functions. In Table~\ref{tab: dynamicsmodel}, we find that although using CD, EMD, or a combination of CD and EMD has similar performance in terms of CD and EMD losses, the Hausdorff distance is significantly improved with the combined loss.   
In Figure~\ref{fig:eval_loss}, we plot the CD and the EMD against the time horizon. We find that although the distance inevitably becomes larger as the model predicts farther into the future, it remains in a reasonable range. This result confirms that it is possible to use these models for manipulation. In Figure~\ref{fig:rigid-motion}, we visualize the predicted particles when rigid motions exist in the sequence. We find that our model is able to predict particles under both rigid motion and non-rigid motion. 

\subsubsection{Manipulation Results} We also evaluate the manipulation results of \model through ablation studies.

\begin{figure}[t]
	\includegraphics[width=\columnwidth]{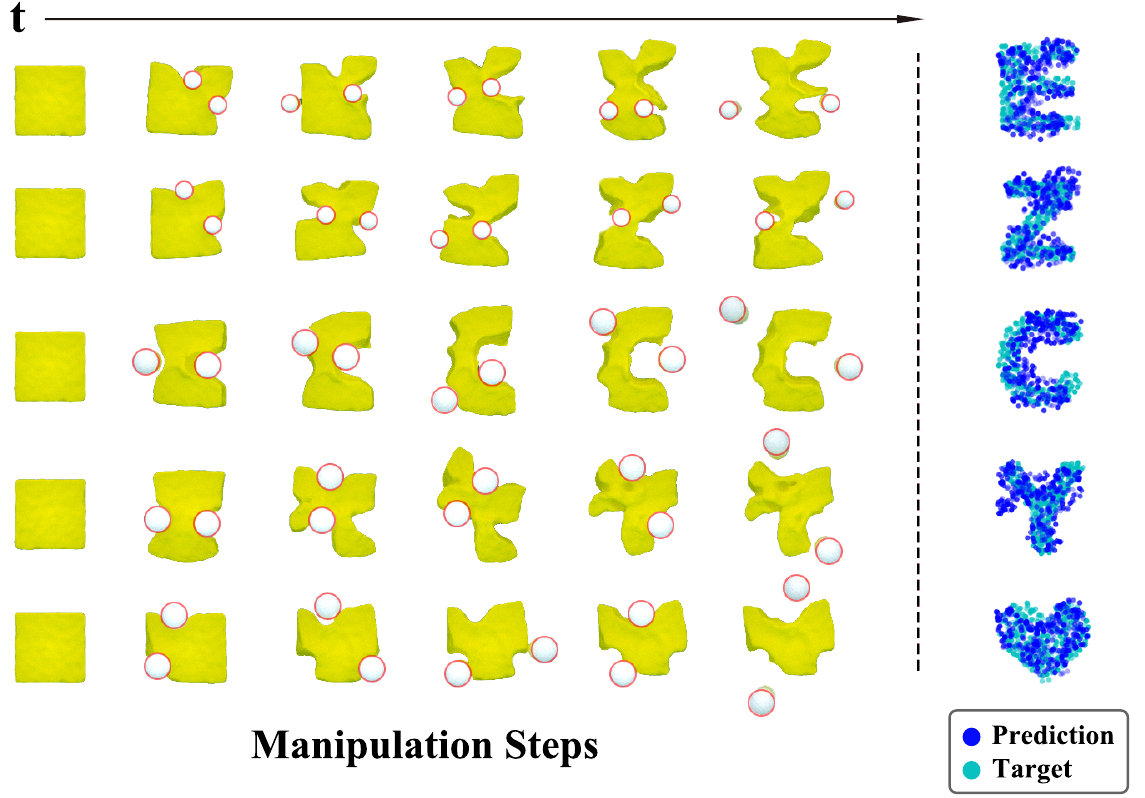}
	\caption{\textbf{Manipulation result in the simulation.} On the left are the manipulation steps. On the right are the result and its overlay with the target point cloud. The cyan point cloud is the target, blue the result.}
	\label{fig:sim-control-result}
\end{figure}

\begin{figure}[t]
	\includegraphics[width=\columnwidth]{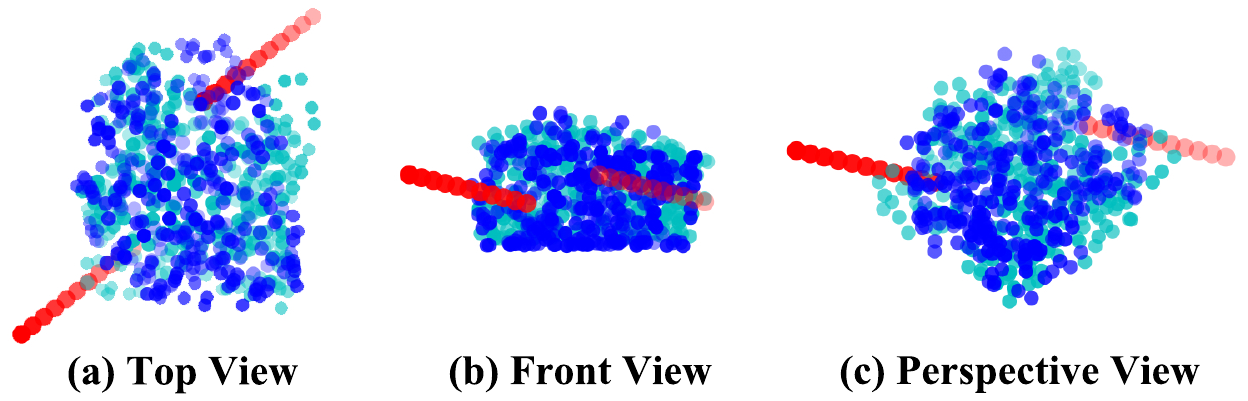}
	\caption{\textbf{The three views of result and target point cloud overlay with a free-moving gripper.} The blue dots are the point cloud of the plasticine, the cyan ones are the target point cloud, and the red dots represent the two fingers of the gripper. }
	\label{fig:grip3d}
\end{figure}

\noindent\textbf{Trajectory optimization methods.} We compare different trajectory optimization methods including random shooting~(RS), gradient-based planning~(GD), gradient-free planning (CEM), and reinforcement learning (model-based Soft Actor-Critic). Specifically, for the reinforcement learning baseline, we use the same action space as in other planning methods. The state-space consists of the position of all the particles and also the gripper. The reward function is computed from the difference in CD or EMD after each grip. For training, we use a discount factor of 0.99 and a learning rate of 0.001 with the Adam optimizer.
We use 2-layer MLPs with 256 hidden units and ReLU activation for both the policy and critic
models. We initially collected $50$ episodes of warm-up data before training. The replay buffer size is 1e6 and the target smooth coefficient is $0.005$. As shown in Figure~\ref{fig:opt}, we find that gradient-based optimization with the learned model outperforms all other methods. We attribute this to the strong optimization power of gradient-based optimization in the simplified action space. We note that the \textit{limited-time BFGS} optimizer provides the best results in practice. 

In Figure~\ref{fig:sim-control-result}, we visualize the procedure of manipulating the object towards the target shape using a gradient-based method. We find that the agent can handle various challenges such as small grooves in the letter `E' and asymmetry in the letter `Z'. This demonstrates that the method can leverage the GNN-based dynamics model for effective manipulation under the MPC framework. More visualized results can be found in the video material.

\begin{table}[t] 
\caption{Results for tool selection averaged over all the target shapes. }
\centering
\begin{tabular}{lcc}
\toprule
Methods         & CD$\downarrow$ & EMD$\downarrow$  \\ 
\midrule
w/ Tool Selection        & \textbf{0.0408} $\pm$ 0.005  & \textbf{0.0337} $\pm$ 0.007  \\
w/o Tool Selection     & 0.0409 $\pm$ 0.005  & 0.0345 $\pm$ 0.007  \\
\bottomrule
\end{tabular}
\label{tab: tool}
\end{table}

\begin{figure}[t!]
	\includegraphics[width=\columnwidth]{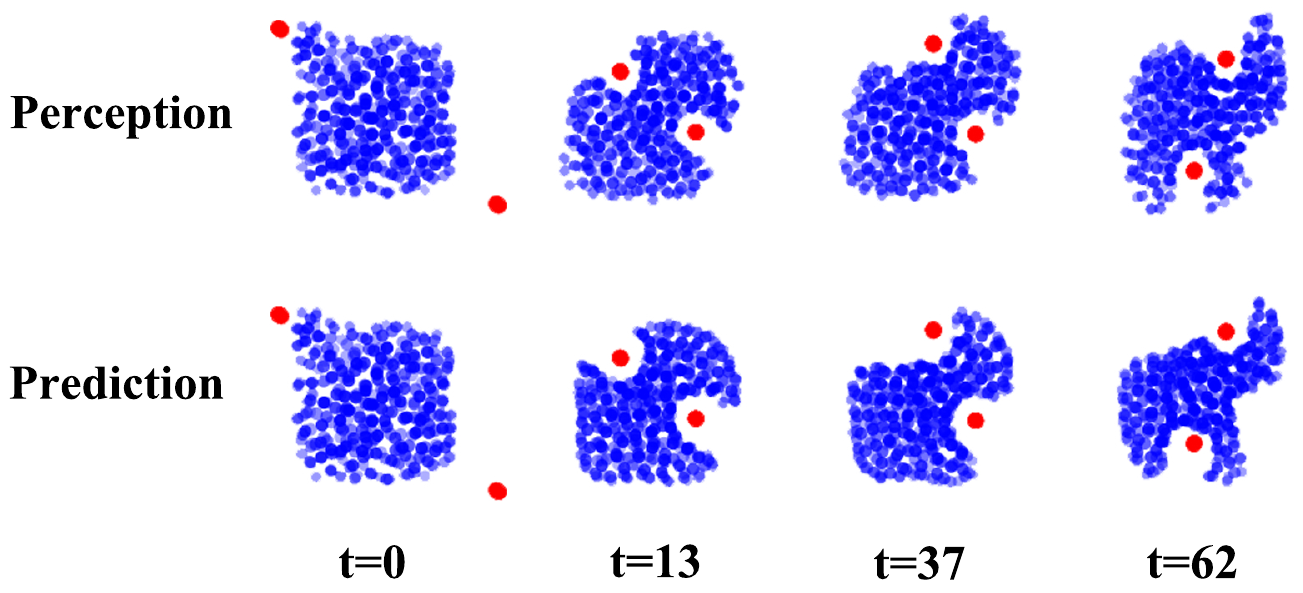}
	\caption{\textbf{Dynamics prediction of the plasticine in the real world.} The first row shows the sampled particles from the actual visual observation. The second row is the open-loop prediction of the particle distributions using our learned dynamics from just 10 minutes of real-world interaction data. Red dots represent the gripper's location.}
	\label{fig:real-dy}
\end{figure}

\noindent\textbf{Complex action space.} We evaluate our method when all DoFs are actuated for the robot gripper, namely \textit{Shaping with 7 DOF gripper}. In Figure \ref{fig:grip3d}, we show the top view, front view, and perspective view of the overlaid result and the target point cloud. We find that the proposed method can learn to rotate the gripper in order to achieve the target shape. This experiment shows the potential of the proposed method to scale to more complex action spaces.

\noindent\textbf{Tool selection.} We evaluate our method when two grippers with different sizes can be selected for each grip. In Table~\ref{tab: tool}, we find that when more tool choices are provided, the performance can be further improved.

\begin{figure*}[t]
	\includegraphics[width=2\columnwidth]{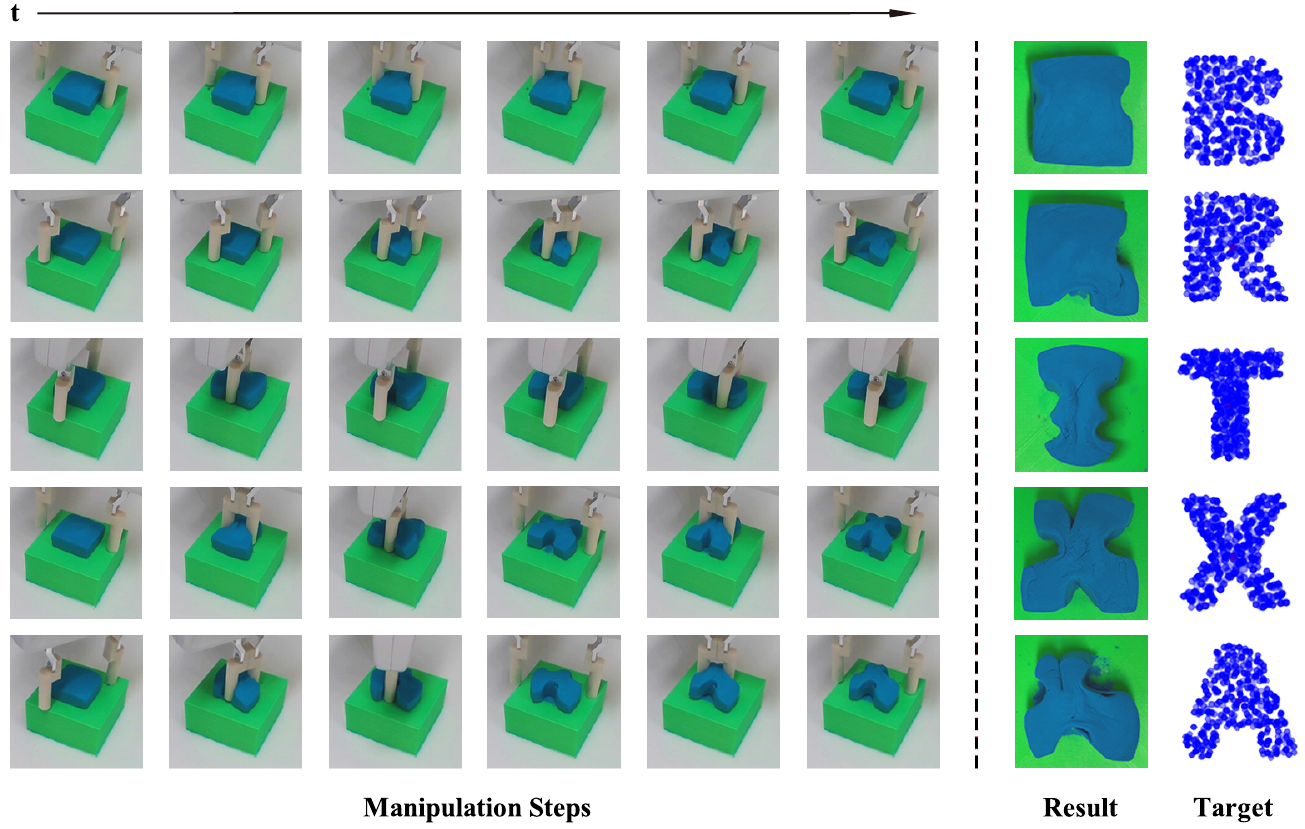}
	\caption{\textbf{Manipulation results on a real robot.} On the left are the shaping steps. On the right are the results and corresponding target point clouds. We want to emphasize that our model is learned purely from offline data collected via random interactions~(10~minutes); thus, the target shapes have never been seen during training. Yet our pipeline can still achieve these targets with reasonable accuracy.}
	\label{fig:real-control-result}
\end{figure*}

\subsection{Learning Real-World Manipulation of Deformable Objects}
We now present an evaluation of our method on a real robot. We first evaluate the quality of the dynamics model with only 10 minutes of real-world data. In Figure~\ref{fig:real-dy}, we visualize the comparison between the output from the trained model and the sampled data. Our method is able to model the dynamics even with limited training data. We then show that the proposed method is able to manipulate the plasticine to shapes that are unseen in the training data. Example trajectories of the robot manipulating the plasticine are shown in Figure~\ref{fig:real-control-result}. The method successfully identifies the asymmetry in the target shape `B' by putting the finger closer to one side of the plasticine at the beginning of the grip. For more complex shapes such as the letter `A', the method also seems to creatively discover a solution that roughly achieves the target shape. These results illustrate that, although the task is very challenging, our method is able to perform well with a small amount of training data.

\subsection{Comparing with Other Dynamics Models in the Real World}
\begin{figure}[t]
	\includegraphics[width=\columnwidth]{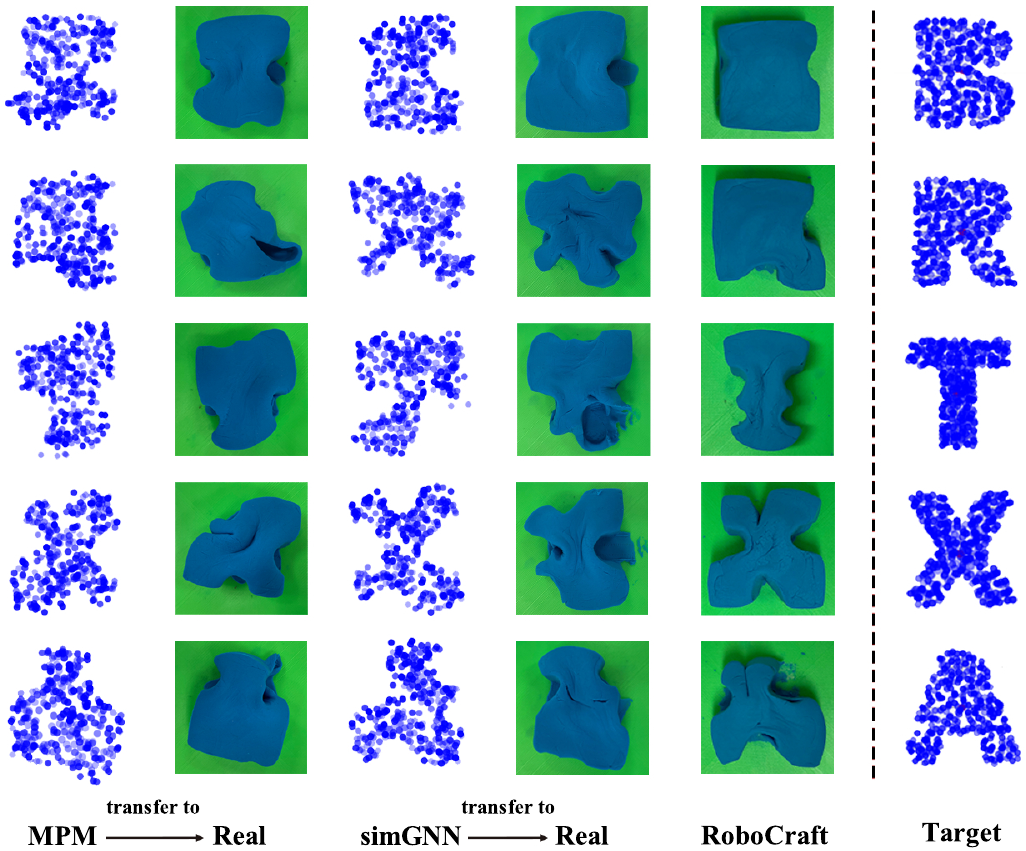}
	\caption{\textbf{\model outperforms methods that use Material Point Method~(MPM) or GNN-based model trained in simulation~(simGNN) as the dynamics model.} The first two columns show the outcomes of transferring control signals from MPM models into the real world. The third and fourth columns show the outcomes of transferring the simGNN model into the real world. We list the outcomes of \model as well as the target shapes for comparison.}
	\label{fig:sim2real}
\end{figure}

\model uses a GNN-based dynamics model learned from real-world data. We compare its efficacy with two other widely adopted options: 1) a physics simulator based on Material Point Method~(MPM) with manually selected parameters and 2) a GNN learned in the simulator with simulation data. We follow previous  work~\citep{huang2021plasticinelab} to create the MPM dynamics model. As shown in Figure~\ref{fig:sim2real}, \model outperforms both baselines. We attribute this to the fact that \model learns from real-world data while other models suffer from the large domain gap between simulation and reality. 

\subsection{Generalizing to Novel Initial Shapes and Material Types}
\begin{figure}[t]
	\includegraphics[width=\columnwidth]{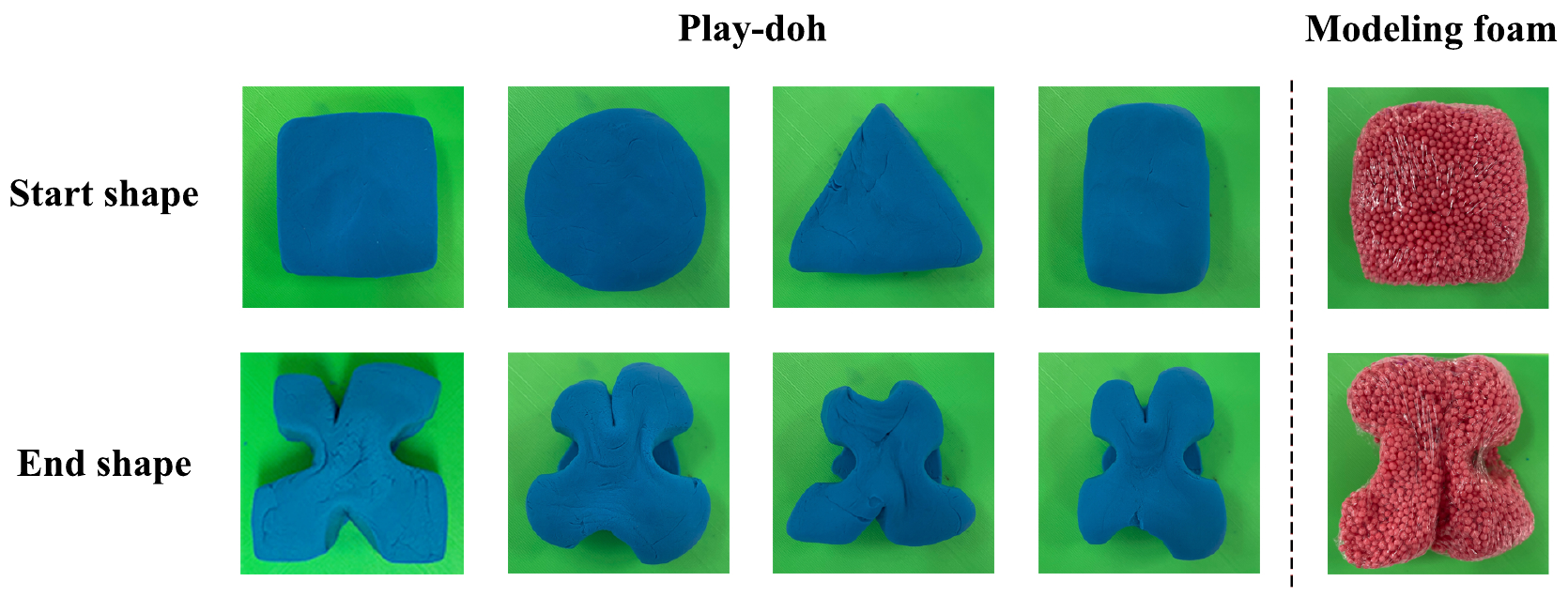}
	\caption{\textbf{\model generalizes to different initial shapes and material types.} Left: \model is applied on $3$ different initial shapes including circle, triangle, and rectangle. Right: \model is applied on modeling foam.}
	\label{fig:generalization}
\end{figure}

We test the generalization ability of \model by applying the framework to different initial object shapes and material types~(modeling foam). As shown on the left of Figure~\ref{fig:generalization}, we find that \model can manipulate the objects with a circle, triangle, or rectangle as their initial shapes into our target shape `X'. We also display the result of the original square shape in this figure for comparison. In the right side of Figure~\ref{fig:generalization}, we test whether the dynamics model of \model which is trained on plasticine can be used to manipulate modeling foam. We find that it can also roughly make an `X' shape without retraining. The results illustrate the potential of \model to generalize to various unseen scenarios.    

\subsection{Results on Comparing with Human Performance}

\begin{table}[t] 
\caption{Results of human subjects and the robot in the simulator. Numbers are averaged over all the tested shapes.}
\centering
\begin{tabular}{lcc}
\toprule
Methods         & CD$\downarrow$ & EMD$\downarrow$  \\ 
\midrule
Human Subjects        & 0.0655 $\pm$ 0.025  & 0.0661 $\pm$ 0.023  \\
\model (ours)     & \textbf{0.0359} $\pm$ 0.007  & \textbf{0.0340} $\pm$ 0.005 \\
\bottomrule
\end{tabular}
\label{tab: human}
\end{table}

\begin{figure}[t]
	\includegraphics[width=\columnwidth]{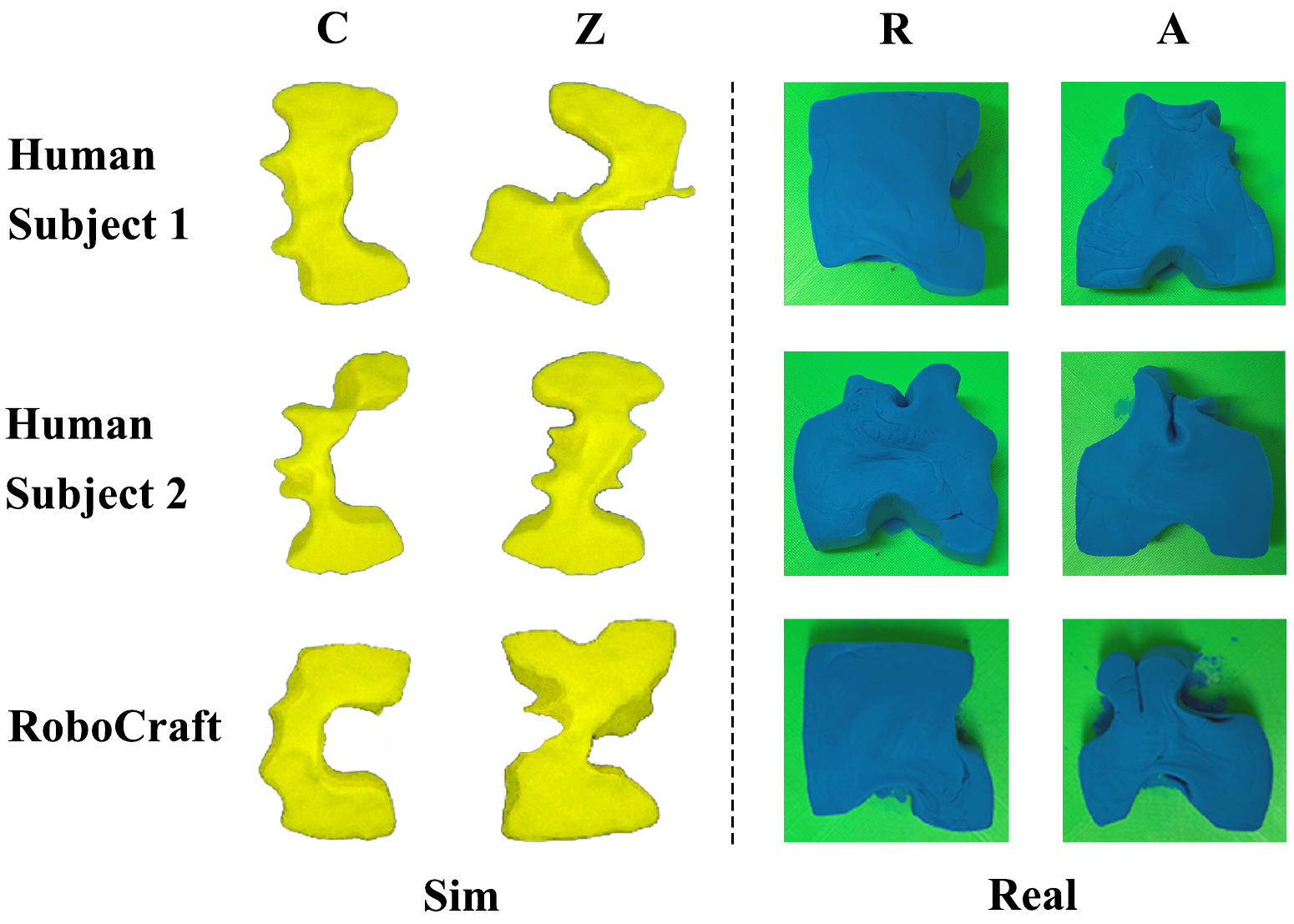}
	\caption{\textbf{Shaping results by amateur humans and \model.} Results in the first two rows are from human subjects. The results in the third row are from the robot. The left two columns are results in the simulation. The right two columns are the results in the real world.}
	\label{fig:human-control-result}
\end{figure}

We invited four amateur humans to perform the same task with the robot gripper in both simulation and real-world settings. While humans are not trained to manipulate plasticine, they usually have strong intuitive understandings of the dynamics of plasticine. Each user was asked to shape three pieces of plasticine in each domain. For the simulation experiments, we provided a successful manipulation video as an example for the users and allowed two trials for each shape since the dynamics in the simulation were also new to each human. The human subjects controlled the gripper using the keyboard to shape the plasticine to `C', `E', `Y', `Z', and heart. In the real world, we allowed the users to play with the plasticine for one minute before resetting it to the starting shape for the experiment. After that, human users controlled the robot arm and gripper manually in the guiding mode to shape the plasticine to `A', `B', `R', `T', and `X'. In Table~\ref{tab: human}, we show the average distance over all the target shapes from four users, in comparison with that of our agent. Empirical evidence suggests that the proposed tasks are challenging for both manipulation algorithms and untrained human subjects alike. We also find that \model is comparable to or stronger than amateur humans on the tested tasks. One observation is that \model outperforms humans in the distance metrics. However, the visualized human results are recognizable even when the distances are high, suggesting that better evaluation metrics are desired. In Figure~\ref{fig:human-control-result}, we show the outcome from both human users and the robot for comparison.

\section{Conclusion} 
\label{sec:conclusion}
We have proposed the first model-based framework that manipulates elasto-plastic objects to complex unseen target shapes both in simulation and on a real robot, with only $10$ minutes of robot exploration time. Real robot experiments demonstrate the high efficacy of the proposed system, with the control algorithm outperforming amateur human users. Furthermore, we explore the possibility of applying the proposed method to harder tasks with complex action spaces, mixtures of motion modes, and multiple tools. 

\section{Acknowledgement}
We thank Samuel Clarke for helpful discussions and hardware support. We also thank Stephen Tian for his careful proofreading and writing suggestions. This work is in part supported by the Stanford Institute for Human-Centered AI (HAI), the Samsung Global Research Outreach (GRO) Program, the Toyota Research Institute (TRI), and Amazon, Autodesk, Salesforce, and Bosch.

\bibliographystyle{plainnat}
\bibliography{references}

\begin{thebibliography}{64}
\providecommand{\natexlab}[1]{#1}
\providecommand{\url}[1]{\texttt{#1}}
\expandafter\ifx\csname urlstyle\endcsname\relax
  \providecommand{\doi}[1]{doi: #1}\else
  \providecommand{\doi}{doi: \begingroup \urlstyle{rm}\Url}\fi

\bibitem[Antonova et~al.(2021{\natexlab{a}})Antonova, Shi, Yin, Weng, and
  Jensfelt]{antonova2021dynamic}
Rika Antonova, Peiyang Shi, Hang Yin, Zehang Weng, and Danica~Kragic Jensfelt.
\newblock Dynamic environments with deformable objects.
\newblock In \emph{Advances in Neural Information Processing Systems (NeurIPS)
  Datasets and Benchmarks Track}, 2021{\natexlab{a}}.

\bibitem[Antonova et~al.(2021{\natexlab{b}})Antonova, Varava, Shi, Carvalho,
  and Kragic]{antonova2021sequential}
Rika Antonova, Anastasia Varava, Peiyang Shi, J~Frederico Carvalho, and Danica
  Kragic.
\newblock Sequential topological representations for predictive models of
  deformable objects.
\newblock In \emph{Learning for Dynamics and Control}, pages 348--360. PMLR,
  2021{\natexlab{b}}.

\bibitem[Antonova et~al.(2022)Antonova, Yang, Sundaresan, Fox, Ramos, and
  Bohg]{antonova2021bayesian}
Rika Antonova, Jingyun Yang, Priya Sundaresan, Dieter Fox, Fabio Ramos, and
  Jeannette Bohg.
\newblock A bayesian treatment of real-to-sim for deformable object
  manipulation.
\newblock \emph{IEEE Robotics and Automation Letters}, 7\penalty0 (3):\penalty0
  5819--5826, 2022.

\bibitem[Battaglia et~al.(2016)Battaglia, Pascanu, Lai, Jimenez~Rezende,
  et~al.]{battaglia2016interaction}
Peter Battaglia, Razvan Pascanu, Matthew Lai, Danilo Jimenez~Rezende, et~al.
\newblock Interaction networks for learning about objects, relations and
  physics.
\newblock \emph{Advances in Neural Information Processing Systems (NeurIPS)},
  29, 2016.

\bibitem[Bernardini et~al.(1999)Bernardini, Mittleman, Rushmeier, Silva, and
  Taubin]{bernardini1999ball}
Fausto Bernardini, Joshua Mittleman, Holly Rushmeier, Claudio Silva, and
  Gabriel Taubin.
\newblock The ball-pivoting algorithm for surface reconstruction.
\newblock \emph{IEEE Transactions on Visualization and Computer Graphics},
  5\penalty0 (4):\penalty0 349--359, 1999.

\bibitem[Chang and Pad{\i}r(2020)]{chang2020model}
Peng Chang and Ta{\c{s}}k{\i}n Pad{\i}r.
\newblock Model-based manipulation of linear flexible objects with visual
  curvature feedback.
\newblock In \emph{IEEE/ASME International Conference on Advanced Intelligent
  Mechatronics (AIM)}, pages 1406--1412. IEEE, 2020.

\bibitem[Cherubini et~al.(2020)Cherubini, Ortenzi, Cosgun, Lee, and
  Corke]{cherubini2020model}
Andrea Cherubini, Valerio Ortenzi, Akansel Cosgun, Robert Lee, and Peter Corke.
\newblock Model-free vision-based shaping of deformable plastic materials.
\newblock \emph{The International Journal of Robotics Research}, 39\penalty0
  (14):\penalty0 1739--1759, 2020.

\bibitem[Cretu et~al.(2011)Cretu, Payeur, and Petriu]{cretu2011soft}
Ana-Maria Cretu, Pierre Payeur, and Emil~M Petriu.
\newblock Soft object deformation monitoring and learning for model-based
  robotic hand manipulation.
\newblock \emph{IEEE Transactions on Systems, Man, and Cybernetics, Part B
  (Cybernetics)}, 42\penalty0 (3):\penalty0 740--753, 2011.

\bibitem[Edelsbrunner and M{\"u}cke(1994)]{edelsbrunner1994three}
Herbert Edelsbrunner and Ernst~P M{\"u}cke.
\newblock Three-dimensional alpha shapes.
\newblock \emph{ACM Transactions on Graphics (TOG)}, 13\penalty0 (1):\penalty0
  43--72, 1994.

\bibitem[Fan et~al.(2017)Fan, Su, and Guibas]{fan2017point}
Haoqiang Fan, Hao Su, and Leonidas~J Guibas.
\newblock A point set generation network for 3d object reconstruction from a
  single image.
\newblock In \emph{Proceedings of the IEEE Conference on Computer Vision and
  Pattern Recognition (CVPR)}, pages 605--613, 2017.

\bibitem[Figueroa et~al.(2016)Figueroa, Ureche, and
  Billard]{figueroa2016learning}
Nadia Figueroa, Ana Lucia~Pais Ureche, and Aude Billard.
\newblock Learning complex sequential tasks from demonstration: A pizza dough
  rolling case study.
\newblock In \emph{2016 11th ACM/IEEE International Conference on Human-Robot
  Interaction (HRI)}, pages 611--612. Ieee, 2016.

\bibitem[Fischler and Bolles(1981)]{fischler1981random}
Martin~A Fischler and Robert~C Bolles.
\newblock Random sample consensus: a paradigm for model fitting with
  applications to image analysis and automated cartography.
\newblock \emph{Communications of the ACM}, 24\penalty0 (6):\penalty0 381--395,
  1981.

\bibitem[Fletcher(2013)]{fletcher2013practical}
Roger Fletcher.
\newblock \emph{Practical methods of optimization}.
\newblock John Wiley \& Sons, 2013.

\bibitem[Ganapathi et~al.(2020)Ganapathi, Sundaresan, Thananjeyan, Balakrishna,
  Seita, Grannen, Hwang, Hoque, Gonzalez, Jamali,
  et~al.]{ganapathi2020learning}
Aditya Ganapathi, Priya Sundaresan, Brijen Thananjeyan, Ashwin Balakrishna,
  Daniel Seita, Jennifer Grannen, Minho Hwang, Ryan Hoque, Joseph~E Gonzalez,
  Nawid Jamali, et~al.
\newblock Learning to smooth and fold real fabric using dense object
  descriptors trained on synthetic color images.
\newblock \emph{arXiv preprint arXiv:2003.12698}, 2020.

\bibitem[Ha and Song(2022)]{ha2021flingbot}
Huy Ha and Shuran Song.
\newblock Flingbot: The unreasonable effectiveness of dynamic manipulation for
  cloth unfolding.
\newblock In \emph{Conference on Robot Learning (CoRL)}, pages 24--33. PMLR,
  2022.

\bibitem[Holl et~al.(2019)Holl, Thuerey, and Koltun]{holl2020learning}
Philipp Holl, Nils Thuerey, and Vladlen Koltun.
\newblock Learning to control pdes with differentiable physics.
\newblock In \emph{International Conference on Learning Representations
  (ICLR)}, 2019.

\bibitem[Hoque et~al.(2022)Hoque, Seita, Balakrishna, Ganapathi, Tanwani,
  Jamali, Yamane, Iba, and Goldberg]{fabric_vsf_2021}
Ryan Hoque, Daniel Seita, Ashwin Balakrishna, Aditya Ganapathi, Ajay~Kumar
  Tanwani, Nawid Jamali, Katsu Yamane, Soshi Iba, and Ken Goldberg.
\newblock Visuospatial foresight for physical sequential fabric manipulation.
\newblock \emph{Autonomous Robots}, 46\penalty0 (1):\penalty0 175--199, 2022.

\bibitem[Hu et~al.(2019)Hu, Anderson, Li, Sun, Carr, Ragan-Kelley, and
  Durand]{hu2019difftaichi}
Yuanming Hu, Luke Anderson, Tzu-Mao Li, Qi~Sun, Nathan Carr, Jonathan
  Ragan-Kelley, and Fredo Durand.
\newblock Difftaichi: Differentiable programming for physical simulation.
\newblock In \emph{International Conference on Learning Representations
  (ICLR)}, 2019.

\bibitem[Huang et~al.(2020)Huang, Hu, Du, Zhou, Su, Tenenbaum, and
  Gan]{huang2021plasticinelab}
Zhiao Huang, Yuanming Hu, Tao Du, Siyuan Zhou, Hao Su, Joshua~B Tenenbaum, and
  Chuang Gan.
\newblock Plasticinelab: A soft-body manipulation benchmark with differentiable
  physics.
\newblock In \emph{International Conference on Learning Representations
  (ICLR)}, 2020.

\bibitem[Huttenlocher et~al.(1993)Huttenlocher, Klanderman, and
  Rucklidge]{huttenlocher1993comparing}
Daniel~P Huttenlocher, Gregory~A. Klanderman, and William~J Rucklidge.
\newblock Comparing images using the hausdorff distance.
\newblock \emph{IEEE Transactions on Pattern Analysis and Machine Intelligence
  (TPAMI)}, 15\penalty0 (9):\penalty0 850--863, 1993.

\bibitem[Kazhdan et~al.(2006)Kazhdan, Bolitho, and Hoppe]{kazhdan2006poisson}
Michael Kazhdan, Matthew Bolitho, and Hugues Hoppe.
\newblock Poisson surface reconstruction.
\newblock In \emph{Proceedings of the fourth Eurographics Symposium on Geometry
  Processing}, volume~7, 2006.

\bibitem[Kingma and Ba(2015)]{kingma2015adam}
Diederik~P Kingma and Jimmy Ba.
\newblock Adam: A method for stochastic optimization.
\newblock In \emph{International Conference on Learning Representation (ICLR)},
  2015.

\bibitem[Kurutach et~al.(2018)Kurutach, Tamar, Yang, Russell, and
  Abbeel]{kurutach2018learning}
Thanard Kurutach, Aviv Tamar, Ge~Yang, Stuart~J Russell, and Pieter Abbeel.
\newblock Learning plannable representations with causal infogan.
\newblock \emph{Advances in Neural Information Processing Systems (NeurIPS)},
  31, 2018.

\bibitem[Lee et~al.(2021)Lee, Hamaya, Murooka, Ijiri, and Corke]{lee2021sample}
Robert Lee, Masashi Hamaya, Takayuki Murooka, Yoshihisa Ijiri, and Peter Corke.
\newblock Sample-efficient learning of deformable linear object manipulation in
  the real world through self-supervision.
\newblock \emph{IEEE Robotics and Automation Letters (RA-L)}, 7\penalty0
  (1):\penalty0 573--580, 2021.

\bibitem[Leggard et~al.(2021)Leggard, Schranz, Schweiger, Drgo{\v{n}}a, Falay,
  Gomes, Iosifidis, Abkar, and Larsen]{legaard2021constructing}
Christian~M{\o}ldrup Leggard, Thomas Schranz, Gerald Schweiger, J{\'a}n
  Drgo{\v{n}}a, Basak Falay, Cl{\'a}udio Gomes, Alexandros Iosifidis, Mahdi
  Abkar, and Peter~Gorm Larsen.
\newblock Constructing neural network-based models for simulating dynamical
  systems.
\newblock \emph{ACM Computing Surveys}, 1\penalty0 (1), 2021.

\bibitem[Li et~al.(2022{\natexlab{a}})Li, Huang, Du, Su, Tenenbaum, and
  Gan]{contact}
Sizhe Li, Zhiao Huang, Tao Du, Hao Su, Joshua Tenenbaum, and Chuang Gan.
\newblock Contact points discovery for soft-body manipulations with
  differentiable physics.
\newblock In \emph{International Conference on Learning Representations
  (ICLR)}, 2022{\natexlab{a}}.

\bibitem[Li et~al.(2018)Li, Wu, Tedrake, Tenenbaum, and
  Torralba]{li2018learning}
Yunzhu Li, Jiajun Wu, Russ Tedrake, Joshua~B Tenenbaum, and Antonio Torralba.
\newblock Learning particle dynamics for manipulating rigid bodies, deformable
  objects, and fluids.
\newblock In \emph{International Conference on Learning Representations
  (ICLR)}, 2018.

\bibitem[Li et~al.(2020{\natexlab{a}})Li, Lin, Yi, Bear, Yamins, Wu, Tenenbaum,
  and Torralba]{li2020visual}
Yunzhu Li, Toru Lin, Kexin Yi, Daniel Bear, Daniel Yamins, Jiajun Wu, Joshua
  Tenenbaum, and Antonio Torralba.
\newblock Visual grounding of learned physical models.
\newblock In \emph{International Conference on Machine Learning (ICML)}, pages
  5927--5936. PMLR, 2020{\natexlab{a}}.

\bibitem[Li et~al.(2020{\natexlab{b}})Li, Torralba, Anandkumar, Fox, and
  Garg]{li2020causal}
Yunzhu Li, Antonio Torralba, Anima Anandkumar, Dieter Fox, and Animesh Garg.
\newblock Causal discovery in physical systems from videos.
\newblock \emph{Advances in Neural Information Processing Systems (NeurIPS)},
  33:\penalty0 9180--9192, 2020{\natexlab{b}}.

\bibitem[Li et~al.(2022{\natexlab{b}})Li, Li, Sitzmann, Agrawal, and
  Torralba]{li20213d}
Yunzhu Li, Shuang Li, Vincent Sitzmann, Pulkit Agrawal, and Antonio Torralba.
\newblock 3d neural scene representations for visuomotor control.
\newblock In \emph{Conference on Robot Learning (CoRL)}, pages 112--123. PMLR,
  2022{\natexlab{b}}.

\bibitem[Lin et~al.(2021)Lin, Wang, Olkin, and Held]{lin2020softgym}
Xingyu Lin, Yufei Wang, Jake Olkin, and David Held.
\newblock Softgym: Benchmarking deep reinforcement learning for deformable
  object manipulation.
\newblock In \emph{Conference on Robot Learning (CoRL)}, pages 432--448. PMLR,
  2021.

\bibitem[Lin et~al.(2022{\natexlab{a}})Lin, Huang, Li, Tenenbaum, Held, and
  Gan]{diffskill}
Xingyu Lin, Zhiao Huang, Yunzhu Li, Joshua Tenenbaum, David Held, and Chuang
  Gan.
\newblock Diffskill: Skill abstraction from differentiable physics for
  deformable object manipulations with tools.
\newblock In \emph{International Conference on Learning Representations
  (ICLR)}, 2022{\natexlab{a}}.

\bibitem[Lin et~al.(2022{\natexlab{b}})Lin, Wang, Huang, and
  Held]{lin2021learning}
Xingyu Lin, Yufei Wang, Zixuan Huang, and David Held.
\newblock Learning visible connectivity dynamics for cloth smoothing.
\newblock In \emph{Conference on Robot Learning (CoRL)}, pages 256--266. PMLR,
  2022{\natexlab{b}}.

\bibitem[Luo et~al.(2018)Luo, Xu, Li, Tian, Darrell, and Ma]{xu2018algorithmic}
Yuping Luo, Huazhe Xu, Yuanzhi Li, Yuandong Tian, Trevor Darrell, and Tengyu
  Ma.
\newblock Algorithmic framework for model-based deep reinforcement learning
  with theoretical guarantees.
\newblock In \emph{International Conference on Learning Representations
  (ICLR)}, 2018.

\bibitem[Manuelli et~al.(2021)Manuelli, Li, Florence, and
  Tedrake]{manuelli2020keypoints}
Lucas Manuelli, Yunzhu Li, Pete Florence, and Russ Tedrake.
\newblock Keypoints into the future: Self-supervised correspondence in
  model-based reinforcement learning.
\newblock In \emph{Conference on Robot Learning (CoRL)}, pages 693--710. PMLR,
  2021.

\bibitem[Matas et~al.(2018)Matas, James, and Davison]{matas2018sim}
Jan Matas, Stephen James, and Andrew~J Davison.
\newblock Sim-to-real reinforcement learning for deformable object
  manipulation.
\newblock In \emph{Conference on Robot Learning (CoRL)}, pages 734--743. PMLR,
  2018.

\bibitem[Matl and Bajcsy(2021)]{matl2021deformable}
Carolyn Matl and Ruzena Bajcsy.
\newblock Deformable elasto-plastic object shaping using an elastic hand and
  model-based reinforcement learning.
\newblock In \emph{2021 IEEE/RSJ International Conference on Intelligent Robots
  and Systems (IROS)}, pages 3955--3962. IEEE, 2021.

\bibitem[Miller et~al.(2012)Miller, Van Den~Berg, Fritz, Darrell, Goldberg, and
  Abbeel]{miller2012geometric}
Stephen Miller, Jur Van Den~Berg, Mario Fritz, Trevor Darrell, Ken Goldberg,
  and Pieter Abbeel.
\newblock A geometric approach to robotic laundry folding.
\newblock \emph{The International Journal of Robotics Research (IJRR)},
  31\penalty0 (2):\penalty0 249--267, 2012.

\bibitem[Moenning and Dodgson(2003)]{moenning2003fast}
Carsten Moenning and Neil~A Dodgson.
\newblock Fast marching farthest point sampling.
\newblock Technical report, University of Cambridge, Computer Laboratory, 2003.

\bibitem[Monaghan(1992)]{monaghan1992smoothed}
Joe~J Monaghan.
\newblock Smoothed particle hydrodynamics.
\newblock \emph{Annual Review of Astronomy and Astrophysics}, 30\penalty0
  (1):\penalty0 543--574, 1992.

\bibitem[Mrowca et~al.(2018)Mrowca, Zhuang, Wang, Haber, Fei-Fei, Tenenbaum,
  and Yamins]{mrowca2018flexible}
Damian Mrowca, Chengxu Zhuang, Elias Wang, Nick Haber, Li~F Fei-Fei, Josh
  Tenenbaum, and Daniel~L Yamins.
\newblock Flexible neural representation for physics prediction.
\newblock \emph{Advances in Neural Information Processing Systems (NeurIPS)},
  31, 2018.

\bibitem[M{\"u}ller et~al.(2007)M{\"u}ller, Heidelberger, Hennix, and
  Ratcliff]{muller2007position}
Matthias M{\"u}ller, Bruno Heidelberger, Marcus Hennix, and John Ratcliff.
\newblock Position based dynamics.
\newblock \emph{Journal of Visual Communication and Image Representation},
  18\penalty0 (2):\penalty0 109--118, 2007.

\bibitem[Nadon et~al.(2018)Nadon, Valencia, and Payeur]{nadon2018multi}
F{\'e}lix Nadon, Angel~J Valencia, and Pierre Payeur.
\newblock Multi-modal sensing and robotic manipulation of non-rigid objects: A
  survey.
\newblock \emph{Robotics}, 7\penalty0 (4):\penalty0 74, 2018.

\bibitem[Nagabandi et~al.(2020)Nagabandi, Konolige, Levine, and
  Kumar]{nagabandi2020deep}
Anusha Nagabandi, Kurt Konolige, Sergey Levine, and Vikash Kumar.
\newblock Deep dynamics models for learning dexterous manipulation.
\newblock In \emph{Conference on Robot Learning (CoRL)}, pages 1101--1112.
  PMLR, 2020.

\bibitem[Nair et~al.(2017)Nair, Chen, Agrawal, Isola, Abbeel, Malik, and
  Levine]{nair2017combining}
Ashvin Nair, Dian Chen, Pulkit Agrawal, Phillip Isola, Pieter Abbeel, Jitendra
  Malik, and Sergey Levine.
\newblock Combining self-supervised learning and imitation for vision-based
  rope manipulation.
\newblock In \emph{2017 IEEE International Conference on Robotics and
  Automation (ICRA)}, pages 2146--2153. IEEE, 2017.

\bibitem[Navarro-Alarcon et~al.(2016)Navarro-Alarcon, Yip, Wang, Liu, Zhong,
  Zhang, and Li]{navarro2016automatic}
David Navarro-Alarcon, Hiu~Man Yip, Zerui Wang, Yun-Hui Liu, Fangxun Zhong,
  Tianxue Zhang, and Peng Li.
\newblock Automatic 3-d manipulation of soft objects by robotic arms with an
  adaptive deformation model.
\newblock \emph{IEEE Transactions on Robotics (T-RO)}, 32\penalty0
  (2):\penalty0 429--441, 2016.

\bibitem[Rubner et~al.(1998)Rubner, Tomasi, and Guibas]{rubner1998metric}
Yossi Rubner, Carlo Tomasi, and Leonidas~J Guibas.
\newblock A metric for distributions with applications to image databases.
\newblock In \emph{Sixth International Conference on Computer Vision (IEEE Cat.
  No. 98CH36271)}, pages 59--66. IEEE, 1998.

\bibitem[S{\'a}nchez et~al.(2020)S{\'a}nchez, Wan, and
  Harada]{sanchez2020tethered}
Daniel S{\'a}nchez, Weiwei Wan, and Kensuke Harada.
\newblock Tethered tool manipulation planning with cable maneuvering.
\newblock \emph{IEEE Robotics and Automation Letters (RA-L)}, 5\penalty0
  (2):\penalty0 2777--2784, 2020.

\bibitem[Sanchez et~al.(2018)Sanchez, Corrales, Bouzgarrou, and
  Mezouar]{sanchez2018robotic}
Jose Sanchez, Juan-Antonio Corrales, Belhassen-Chedli Bouzgarrou, and Youcef
  Mezouar.
\newblock Robotic manipulation and sensing of deformable objects in domestic
  and industrial applications: a survey.
\newblock \emph{The International Journal of Robotics Research (IJRR)},
  37\penalty0 (7):\penalty0 688--716, 2018.

\bibitem[Sanchez-Gonzalez et~al.(2020)Sanchez-Gonzalez, Godwin, Pfaff, Ying,
  Leskovec, and Battaglia]{sanchez2020learning}
Alvaro Sanchez-Gonzalez, Jonathan Godwin, Tobias Pfaff, Rex Ying, Jure
  Leskovec, and Peter Battaglia.
\newblock Learning to simulate complex physics with graph networks.
\newblock In \emph{International Conference on Machine Learning (ICML)}, pages
  8459--8468. PMLR, 2020.

\bibitem[Schoenholz et~al.(2018)Schoenholz, Cubuk, and Jax]{schoenholz2018end}
Samuel~S Schoenholz, Ekin~D Cubuk, and MD~Jax.
\newblock End-to-end differentiable, hardware accelerated, molecular dynamics
  in pure python.
\newblock \emph{arXiv preprint arXiv:1912.04232}, 2018.

\bibitem[Schrittwieser et~al.(2020)Schrittwieser, Antonoglou, Hubert, Simonyan,
  Sifre, Schmitt, Guez, Lockhart, Hassabis, Graepel,
  et~al.]{schrittwieser2020mastering}
Julian Schrittwieser, Ioannis Antonoglou, Thomas Hubert, Karen Simonyan,
  Laurent Sifre, Simon Schmitt, Arthur Guez, Edward Lockhart, Demis Hassabis,
  Thore Graepel, et~al.
\newblock Mastering atari, go, chess and shogi by planning with a learned
  model.
\newblock \emph{Nature}, 588\penalty0 (7839):\penalty0 604--609, 2020.

\bibitem[Seita et~al.(2021)Seita, Florence, Tompson, Coumans, Sindhwani,
  Goldberg, and Zeng]{seita2020learning}
Daniel Seita, Pete Florence, Jonathan Tompson, Erwin Coumans, Vikas Sindhwani,
  Ken Goldberg, and Andy Zeng.
\newblock Learning to rearrange deformable cables, fabrics, and bags with
  goal-conditioned transporter networks.
\newblock In \emph{2021 IEEE International Conference on Robotics and
  Automation (ICRA)}, pages 4568--4575. IEEE, 2021.

\bibitem[She et~al.(2021)She, Wang, Dong, Sunil, Rodriguez, and
  Adelson]{she2019cable}
Yu~She, Shaoxiong Wang, Siyuan Dong, Neha Sunil, Alberto Rodriguez, and Edward
  Adelson.
\newblock Cable manipulation with a tactile-reactive gripper.
\newblock \emph{The International Journal of Robotics Research (IJRR)},
  40\penalty0 (12-14):\penalty0 1385--1401, 2021.

\bibitem[Shlomi et~al.(2020)Shlomi, Battaglia, and Vlimant]{shlomi2020graph}
Jonathan Shlomi, Peter Battaglia, and Jean-Roch Vlimant.
\newblock Graph neural networks in particle physics.
\newblock \emph{Machine Learning: Science and Technology}, 2\penalty0
  (2):\penalty0 021001, 2020.

\bibitem[Sulsky et~al.(1995)Sulsky, Zhou, and Schreyer]{sulsky1995application}
Deborah Sulsky, Shi-Jian Zhou, and Howard~L Schreyer.
\newblock Application of a particle-in-cell method to solid mechanics.
\newblock \emph{Computer Physics Communications}, 87\penalty0 (1-2):\penalty0
  236--252, 1995.

\bibitem[Sundaresan et~al.(2020)Sundaresan, Grannen, Thananjeyan, Balakrishna,
  Laskey, Stone, Gonzalez, and Goldberg]{sundaresan2020learning}
Priya Sundaresan, Jennifer Grannen, Brijen Thananjeyan, Ashwin Balakrishna,
  Michael Laskey, Kevin Stone, Joseph~E Gonzalez, and Ken Goldberg.
\newblock Learning rope manipulation policies using dense object descriptors
  trained on synthetic depth data.
\newblock In \emph{2020 IEEE International Conference on Robotics and
  Automation (ICRA)}, pages 9411--9418. IEEE, 2020.

\bibitem[Thananjeyan et~al.(2017)Thananjeyan, Garg, Krishnan, Chen, Miller, and
  Goldberg]{thananjeyan2017multilateral}
Brijen Thananjeyan, Animesh Garg, Sanjay Krishnan, Carolyn Chen, Lauren Miller,
  and Ken Goldberg.
\newblock Multilateral surgical pattern cutting in 2d orthotropic gauze with
  deep reinforcement learning policies for tensioning.
\newblock In \emph{2017 IEEE International Conference on Robotics and
  Automation (ICRA)}, pages 2371--2378. IEEE, 2017.

\bibitem[Wang et~al.(2019)Wang, Kurutach, Liu, Abbeel, and
  Tamar]{wang2019learning}
Angelina Wang, Thanard Kurutach, Kara Liu, Pieter Abbeel, and Aviv Tamar.
\newblock Learning robotic manipulation through visual planning and acting.
\newblock In \emph{Robotics: Science and Systems (RSS)}, 2019.

\bibitem[Wu et~al.(2020)Wu, Yan, Kurutach, Pinto, and Abbeel]{wu2019learning}
Yilin Wu, Wilson Yan, Thanard Kurutach, Lerrel Pinto, and Pieter Abbeel.
\newblock {Learning to Manipulate Deformable Objects without Demonstrations}.
\newblock In \emph{Robotics: Science and Systems (RSS)}, July 2020.
\newblock \doi{10.15607/RSS.2020.XVI.065}.

\bibitem[Yan et~al.(2020)Yan, Zhu, Jin, and Bohg]{yan2020self}
Mengyuan Yan, Yilin Zhu, Ning Jin, and Jeannette Bohg.
\newblock Self-supervised learning of state estimation for manipulating
  deformable linear objects.
\newblock \emph{IEEE Robotics and Automation Letters (RA-L)}, 5\penalty0
  (2):\penalty0 2372--2379, 2020.

\bibitem[Yan et~al.(2021)Yan, Vangipuram, Abbeel, and Pinto]{yan2020learning}
Wilson Yan, Ashwin Vangipuram, Pieter Abbeel, and Lerrel Pinto.
\newblock Learning predictive representations for deformable objects using
  contrastive estimation.
\newblock In \emph{Conference on Robot Learning (CoRL)}, pages 564--574. PMLR,
  2021.

\bibitem[Yin et~al.(2021)Yin, Varava, and Kragic]{yin2021modeling}
Hang Yin, Anastasia Varava, and Danica Kragic.
\newblock Modeling, learning, perception, and control methods for deformable
  object manipulation.
\newblock \emph{Science Robotics}, 6\penalty0 (54), 2021.

\bibitem[Yoshimoto et~al.(2011)Yoshimoto, Higashimori, Tadakuma, and
  Kaneko]{yoshimoto2011active}
Kayo Yoshimoto, Mitsuru Higashimori, Kenjiro Tadakuma, and Makoto Kaneko.
\newblock Active outline shaping of a rheological object based on plastic
  deformation distribution.
\newblock In \emph{2011 IEEE/RSJ International Conference on Intelligent Robots
  and Systems}, pages 1386--1391. IEEE, 2011.

\end{thebibliography}

\end{document}